\def\isarxiv{1}

\def\paperTitle{Structured and Fast Optimization: The Kronecker SGD Algorithm}

\def\paperRTitle{\paperTitle} 

\def\paperAuthor{
Zhao Song\thanks{\texttt{magic.linuxkde@gmail.com}. Simons Institute for the Theory of Computing, UC Berkeley.}
\and
Song Yue\thanks{Northeastern University.}
}

\newcommand{\Zhao}[1]{{\color{red}[Zhao: #1]}}
\ifdefined\isarxiv
\documentclass[11pt]{article}
\usepackage[numbers]{natbib}
\else
\documentclass{article}

\fi

\ifdefined\isarxiv

\usepackage{amsmath}
\usepackage{amsthm}
\usepackage{amssymb}
\usepackage{algorithm}
\usepackage{subfig}
\usepackage{algpseudocode}
\usepackage{graphicx}
\usepackage{grffile}
\usepackage{wrapfig,epsfig}
\usepackage{url}
\usepackage{xcolor}
\usepackage{epstopdf}

\usepackage{bbm}
\usepackage{dsfont}

\else 

\usepackage{microtype}
\usepackage{graphicx}
\usepackage{subcaption}
\usepackage{booktabs} % for professional tables
\usepackage{bbm}
\usepackage{hyperref}

\usepackage{icml2026}

\usepackage{amsmath}
\usepackage{amssymb}
\usepackage{mathtools}
\usepackage{amsthm}

\fi

\ifdefined\isarxiv

\else
\usepackage{algpseudocode}
\fi
 
\allowdisplaybreaks

\ifdefined\isarxiv

\usepackage{tikz}
\usepackage{hyperref}  
\hypersetup{colorlinks=true,citecolor=blue,linkcolor=blue} 
\usetikzlibrary{arrows}
\usepackage[margin=1in]{geometry}

\else
\fi
 
\graphicspath{{./figs/}}

\theoremstyle{plain}
\newtheorem{theorem}{Theorem}[section]
\newtheorem{lemma}[theorem]{Lemma}
\newtheorem{definition}[theorem]{Definition}

\newtheorem{corollary}[theorem]{Corollary}

\newtheorem{fact}[theorem]{Fact}
\newtheorem{remark}[theorem]{Remark}
\newtheorem{claim}[theorem]{Claim}

\newcommand{\wt}{\widetilde}
\newcommand{\ov}{\overline}

\newcommand{\N}{\mathcal{N}}
\newcommand{\R}{\mathbb{R}}

\renewcommand{\d}{\mathrm{d}}

\renewcommand{\varepsilon}{\epsilon}
\renewcommand{\tilde}{\wt}

\newcommand{\fire}{\mathrm{fire}}
\newcommand{\Tmat}{{\cal T}_{\mathrm{mat}}}
\newcommand{\batch}{\mathrm{batch}}

\DeclareMathOperator*{\E}{{\mathbb{E}}}

\DeclareMathOperator{\poly}{poly}

\DeclareMathOperator{\vect}{vec}

\DeclareMathOperator{\dis}{dis}
\DeclareMathOperator{\cts}{cts}

\ifdefined\isarxiv

\else
\icmltitlerunning{\paperRTitle}

\fi

\begin{document}

\ifdefined\isarxiv
%%% The below part is the title and author of ArXiv version.

\date{}
\title{\paperTitle}
\author{\paperAuthor}

\else
%%% The below part is the title and author of ICML version.

\twocolumn[
  \icmltitle{\paperTitle}


  \icmlsetsymbol{equal}{*}

  \begin{icmlauthorlist}
    \icmlauthor{Firstname1 Lastname1}{equal,yyy}
    \icmlauthor{Firstname2 Lastname2}{equal,yyy,comp}
    \icmlauthor{Firstname3 Lastname3}{comp}
    \icmlauthor{Firstname4 Lastname4}{sch}
    \icmlauthor{Firstname5 Lastname5}{yyy}
    \icmlauthor{Firstname6 Lastname6}{sch,yyy,comp}
    \icmlauthor{Firstname7 Lastname7}{comp}
    \icmlauthor{Firstname8 Lastname8}{sch}
    \icmlauthor{Firstname8 Lastname8}{yyy,comp}
  \end{icmlauthorlist}

  \icmlaffiliation{yyy}{Department of XXX, University of YYY, Location, Country}
  \icmlaffiliation{comp}{Company Name, Location, Country}
  \icmlaffiliation{sch}{School of ZZZ, Institute of WWW, Location, Country}

  \icmlcorrespondingauthor{Firstname1 Lastname1}{first1.last1@xxx.edu}
  \icmlcorrespondingauthor{Firstname2 Lastname2}{first2.last2@www.uk}

  \icmlkeywords{Machine Learning, ICML}

  \vskip 0.3in
]

\printAffiliationsAndNotice{} 

\fi

\ifdefined\isarxiv
\begin{titlepage}
  \maketitle
  \begin{abstract}
Stochastic gradient descent (SGD) now acts as a fundamental part of optimization in current machine learning. Meanwhile, deep learning architectures have shown outstanding performance in a wide range of fields, such as natural language processing, bioinformatics, and computer vision. Nevertheless, as the parameter size $d$ increases, these models encounter serious efficiency challenges. Previous studies show that the per step calculation expense scales linearly with the input size $d$. To mitigate this, our paper explores inherent patterns, such as Kronecker products within the training examples. We consider input data points that
can be represented as tensor products of lower-
dimensional vectors. We introduce a novel stochastic optimization method where the computational load for every update scales sublinearly with $d$, assuming moderate structural properties of the inputs. We believe our research is the first work achieving this result, representing a significant step forward for efficient deep learning optimization. Our theoretical findings are supported by a formal theorem, demonstrating that the proposed algorithm can train a two-layer fully connected neural network with a per-iteration cost independent of $d$.

%in time $o(m)\cdot \#\mathrm{data}$, which is independent of the data dimensionality $d$, and sublinear in network width $m$. %$n$ is the number of input data points.  

  \end{abstract}
  \thispagestyle{empty}
\end{titlepage}

%{\hypersetup{linkcolor=black}
%\tableofcontents
%}
\newpage

\else

\begin{abstract}

\end{abstract}

\fi

%%% The part below is the main body and reference

\section{Introduction}
\iffalse
Deep neural networks have achieved great success in many fields, e.g., computer vision~\cite{lbbh98,ksh12,sljsraevr15,sz15,hzrs16}, natural language processing~\cite{cwbkkk11,dclt18,pnigclz18,rnss18,ydyc+19}, and bioinformatics~\cite{mly17}, just to name a few. In spite of the excellent performances in a variety of applications, the deep neural networks have brought intensive computation and occupied large storage with the growth of layers. For example, the \textit{ResNet} proposed by~\cite{hzrs16} has $152$ layers and the parameters of VGG-$16$~\cite{sz15} take $552$MB memory~\cite{hmd15}. In order to get around these problems, a lot of relevant approaches have been proposed. In terms of the storage issue, \cite{hmd15,hptd15} compressed the deep networks significantly without loss of accuracy by pruning redundant connections between layers, and deployed the compressed networks on embedded systems. For the intensive computation issue, researchers have focused on reducing training time for deep networks. Total training time involves the number of iterations and the time cost per iteration. We focus on the latter aspect in our work.
%resnet use imagenet dataset;vgg 16 use ILSVRC dataset
\fi
Deep neural networks have achieved great success in many fields, e.g., computer vision~\cite{lbbh98,ksh12,sljsraevr15,sz15,hzrs16}, natural language processing~\cite{cwbkkk11,dclt19,pnigclz18,rnss18,ydyc+19}, and bioinformatics~\cite{mly17}, just to name a few. In spite of the excellent performances in a variety of applications, the deep neural networks have brought intensive computation, and the increasing dimension of the data will affect the performance and computation cost. For example, classification models such as \textit{ResNet} proposed by~\cite{hzrs16} and VGG-16~\cite{sz15} use $224 \times 224$ pixel figures, and generative models such as Stable Diffusion~\cite{rbleo22} use $512\times512, 768\times768,$ and $1024\times1024$ pixel figures. In order to get around these problems, a lot of relevant approaches have been proposed. In terms of the storage issue, \cite{hmd15,hptd15} compressed the deep networks significantly without loss of accuracy by pruning redundant connections between layers, and deployed the compressed networks on embedded systems. For the intensive computation issue, researchers have focused on reducing training time for deep networks. Total training time involves the number of iterations and the time cost per iteration in SGD algorithm. We focus on the latter aspect in our work.

In order to execute faster computation in each iteration, one natural choice is to utilize some high-dimensional data structures which can query points in some geometric regions efficiently. The first kind of methods are based on locality-sensitive hashing (LSH)~\cite{im98} which returns a point from a set that is closest to a given query point under some metric (e.g., $\ell_p$ norm). \cite{clpxldssr20} built an end-to-end LSH framework MONGOOSE to train neural network efficiently via a modified learnable version of data-dependent LSH. \cite{cmfts20} proposed SLIDE that significantly reduces the computations in both training and inference stages by taking advantage of nearest neighbor search based on LSH. \cite{sx21} proposed a unified framework LSH-\textsc{SmiLe} that scales up both forward simulation and backward learning by the locality of partial differential equations update. The second kind of methods utilize the data structures on space partitioning, including $k$-$d$ tree~\cite{Bentley75,chan19}, Quadtree~\cite{fb74}, and partition tree~\cite{agarwal92,matousek92a,matousek92b,ac09,chan12}, etc. \cite{syz21} employed the Half-Space Reporting (HSR) data structures~\cite{agarwal92}, which are able to return all the points having large inner products and support data updates, and improved the time complexity of each iteration in training neural networks to sublinear in network width. 

The above-mentioned works have tried accelerating the training time of deep networks from the perspective of data structures. In this paper, we try doing that from the perspective of input data. A natural question to ask is that 

{\it Is there some mild assumption on the input data, so that each iteration only takes sublinear time in the data dimension in training neural networks?}

In this work, we answer this question positively. To the best of our knowledge, all the previous work needs to pay linear in data dimension $d$ at each iteration \cite{ll18,dzps19,als19_dnn,als19_rnn,dllwz19,sy19,syz21}. This is the first work that achieves the cost per iteration independent of dimensionality $d$.

About the Kronecker structure, we are motivated from the phenomenon that the training data often have a variety of features extracted from different methods or domains. To enhance robustness and discrimination, researchers combine various features into one holistic feature using tensor products before training. Specifically, given two vectors \(u \in \mathbb{R}^{d_1}\) and \(v \in \mathbb{R}^{d_2}\) representing different features, the tensor product \(u \otimes v\) gives a \(d_1 \times d_2\) matrix, which can be vectorized to a \((d_1 \times d_2)\)-dimensional vector.
In bioinformatics field~\cite{bn05}, the fusion of different features of proteins can help us analyze their characteristics effectively. \cite{shl09} employed tensor product feature space to model interactions between feature sets in different domains and proposed two methods to circumvent the feature selection problem in the tensor product feature space. For click-through rate prediction~\cite{jzc+16,nmd+19}, the accuracy can be improved by fusion of two features. In computer vision, \cite{zbll12} combined three features HOG~\cite{dt05}, LBP~\cite{opm02}, and Haar-like~\cite{byb10} by tensor products and applied the new feature to visual tracking. Apart from the above-mentioned applications, the Kronecker structure has also been widely applied to deep neural networks~\cite{gao2020kronecker,jagtap2022deep,feng2024deep,patro2023application}. 

\iffalse
\paragraph{Our Results}
We try improving the cost of per iteration when the input data has some special structures. Assume that the input data points satisfy that for any $i\in[n]$, $x_i = \vect(\ov{x}_i\ov{x}_i^\top) \in \R^d$ with $\ov{x}_i\in\R^{\sqrt{d}}$. For this setting, we have the following theorem, which is our main contribution.  
\begin{theorem}[Informal version of Theorem~\ref{thm:main_formal}]\label{thm:main_informal}
Given $n$ training samples $\{(x_i, y_i)\}_{i=1}^{n}$ such that for each $i\in[n]$, $x_i=\vect(\ov{x}_i\ov{x}_i^\top)\in\R^d$, there exists a stochastic gradient descent algorithm that can train a two-layer fully connected neural network such that each iteration takes time $o(m)\cdot n$, where $m$ is the width of the neural network, that is independent of the data dimension $d$. 
\end{theorem}
The conclusion also holds for the general case: for each $i\in[n]$, $x_i=b_i\otimes a_i\in\R^d$ with $a_i\in\R^p$, $b_i\in\R^q$ and $p,q=O(\sqrt{d})$.
\fi

\paragraph{Our Results}
We try improving the cost of per iteration when the input data has some special structures. Assume that the input data points satisfy Kronecker property. For this setting, we have the following theorem, which is our main contribution.  
\begin{theorem}[Informal version of Theorem~\ref{thm:main_formal}]\label{thm:main_informal}
Given $n$ training samples $\{(x_i, y_i)\}_{i=1}^{n}$ such that for each $i\in[n]$, $x_i\in\R^d$ satisfies Kronecker property, there exists a stochastic gradient descent algorithm that can train a two-layer shifted ReLU activated neural network with $m$ neurons in the hidden layer such that each iteration takes time $o(m)\cdot n$, where $m$ is the width of the neural network, that is independent of the data dimension $d$. 
\end{theorem}
The conclusion also holds for the general case: for each $i\in[n]$, $x_i=b_i\otimes a_i\in\R^d$ with $a_i\in\R^p$, $b_i\in\R^q$ and $p,q=O(\sqrt{d})$.The details of Kronecker property are in Section~\ref{sec:technical_overview}. The above result shows that our algorithm could train neural network with high dimensional input data  faster than previous works under mild assumptions. %%% Section 1. Introduction
\section{Related Work}
\paragraph{Kernel Matrix.} Let $X:=[x_1, \cdots, x_n]\in\R^{d\times n}$, then the Gram matrix $G\in\R^{n\times n}$ of the $n$ columns of $X$ satisfies that $G_{i,j}=x_i^\top x_j$, i.e., $G=X^\top X$. The Gram matrix $K\in\R^{n\times n}$ with respect to the $n$ columns of $X$ such that $K_{i,j}=k(x_i, x_j)$ is called a kernel matrix, where $k$ is referred to as a kernel function.

\cite{daniely17} showed that in polynomial time, the stochastic gradient descent algorithm can learn a function which is competitive with the best function in the conjugate kernel space %\Junwei{What's conjugate kernel space?}\Zhao{I also don't what is this.} 
of the network, and established connection between neural networks and kernel methods. \cite{jgh18} proved that for a multi-layer fully connected neural network, if the weight matrix of each layer has infinite width, then the convergence of gradient descent method can be described by the Neural Tangent Kernel (NTK). \cite{dzps19} researched a two-layer neural network with ReLU activation function, which is not smooth, and proved that the Gram matrix, which is the kernel matrix in~\cite{jgh18}, keeps stable in infinite training time. 

\textbf{Convergence of Neural Network.} There have been two lines of work proving the convergence of neural networks: the first is based on the assumption that the input data are from Gaussian distribution; the other follows the NTK regime~\cite{jgh18,ll18,dzps19,als19_dnn,als19_rnn,gll+24b,swl22,swl24}. In~\cite{jgh18}, the NTK is first proposed and is central to characterize the generalization features of neural networks. Moreover, it is proven that the convergence of training is related to the positive-definiteness of the limiting NTK. \cite{ll18} observed that in the training of a two-layer fully connected neural network, a fraction of neurons are not activated over iterations, i.e., $w_r(t)^\top x_i\le\tau$, where $r\in[m]$ and $\tau \in \R$ is the threshold of the shifted ReLU activation function. Based on this observation, \cite{ll18} obtained the convergence rate by using stochastic gradient descent to optimize the cross-entropy loss function. However, the network width $m$ depends on $\poly(1/\epsilon)$, %\Junwei{why when the accuracy is low, m (width of neural network) is high? Is it because of overfitting?}\Zhao{I don't know.}
where $\epsilon$ is the desired accuracy, and approaches infinity when $\epsilon$ approaches $0$. In~\cite{dzps19}, the lower bound of $m$ is improved to $\poly(n, 1/\rho, \log(m/\rho))$, where $\rho$ is the probability parameter, by setting the amount of over-parameterization to be independent of $\epsilon$. 

\paragraph{Roadmap} This paper is organized as follows: In Section~\ref{sec:notation}, we define notations used in the paper. In Section~\ref{sec:problem}, we illustrated the gradient descent formula for shifted ReLU activated neural network. In Section~\ref{sec:technical_overview}, we introduced the functionality of proposed data structure. For Section~\ref{sec:efficient_sgd}, we demonstrated the training algorithm. In Section~\ref{sec:convergence_complextiy}, we introduced  the convergence theorem and complexity analysis for our algorithm. Then we provide a discussion on our technical novelty and summarize our contributions in Section~\ref{sec:discussion}. 
% Finally, we summarize our results and contributions in Section~\ref{sec:conclusion}. 
\section{Notation}\label{sec:notation}
% For a positive integer $n$, let $[n]$ represent the set $\{1, 2, \cdots, n\}$. 
Let $\vect(\cdot)$ denote the \textit{vectorization} operator. Specifically, for a matrix $A=[a_1, \cdots, a_d]\in\R^{d\times d}$, $\vect(A)=[a_1^\top, \cdots, a_d^\top]^\top\in\R^{d^2}$. For the $\vect(\cdot)$ operator, let $\vect^{-1}(\cdot)$ be its inverse such that $\vect^{-1}(\vect(A))=A$. For $n \in \R$,let $\poly (n)$ denote a polynomial function of $n$. For a matrix $A\in\R^{d\times n}$ and a subset $S\subset [n]$, $A_{i,j}$ is the entry of $A$ at the $i$-th row and the $j$-th column, and $A_{:, S}$ represents the matrix whose columns correspond to the columns of $A$ indexed by the set $S$. Similarly, for a vector $x\in\R^n$, $x_{S}$ is a vector whose entries correspond to the entries in $x$ indexed by the set $S$. 
Let $\|\cdot\|_2$ and $\|\cdot\|_F$ represent the $\ell_2$ norm and Frobenius norm respectively. The symbol $\mathbbm{1}(\cdot)$ represents the indicator function. For a positive integer $d$, $I_d$ denotes the $d\times d$ identity matrix. For two vectors $a, b\in\R^n$, let $a\odot b\in\R^n$ represent the entry-wise product of $a$ and $b$. For any two matrices $A$ and $B$, $A\otimes B$ represents the Kronecker product of $A$ and $B$. For a random variable $X$, let $\E[X]$ denote the expectation of $X$. The symbol $\Pr[\cdot]$ represents probability. 
% Due to space constraints, we move some notation to Section~\ref{sec:notations}.

\section{Problem Formulation}\label{sec:problem}
Our problem formulation is similar to that of~\cite{dzps19,sy19,syz21}. Define the shifted ReLU function to be  $ \phi_{\tau}(x):=\max\{x-\tau, 0\}$, 
where $x, \tau\in\R$ and $\tau\ge 0$ is the threshold. We consider a two-layer shifted ReLU activated neural network with $m$ neurons in the hidden layer 
$
f (W,a,x) := \frac{1}{ \sqrt{m} } \sum_{r=1}^m a_r\cdot\phi_\tau  ( w_r^\top x  ),
$
where $x \in \R^d$ is the input, $W=\{w_1, \cdots, w_m\}\subset\R^d$ are weight vectors in the first layer, and $a_1, \cdots, a_m \in \R$ are weights in the second layer. 
While shifted ReLU may not be commonly used in all applications, it offers theoretical advantages that help in our convergence proofs. We chose it for its mathematical properties that facilitate our analysis,and many other works such as~\cite{syz21,als+23,qsy24,gqsw24} also use shifted ReLU. Additionally, in our settings,$\tau$ could be a very small positive number, which is close to classical ReLU function. The assumption regarding control of neuron activation is important for ensuring convergence and efficiency of the proposed method. 
For simplicity, we only optimize the $m$ weight vectors $w_1, \cdots, w_m$ without training $a$. Then for each $r\in [m]$,
we have 
$
\frac{\partial f(W,a,x)}{\partial w_r}=\frac{1}{ \sqrt{m} } a_r\cdot\mathbbm{1} ( w_r^\top x>\tau  )\cdot x.    
$
Given $n$ training samples $\{(x_i, y_i)\}_{i=1}^n$ with $x_i\in\R^d$ and $y_i\in\R$ for each $i\in[n]$, the objective function $L(W)$ is defined by 
$
    L (W) := \frac{1}{2} \sum_{i=1}^n (f(W,a,x_i)-y_i)^2 .
$

Additionally, for a specific batch $S\subseteq [n]$, the objective function denoted as $L(W, S)$ is defined to be 

$
    L(W, S):= \frac{n}{|S|}\cdot\frac{1}{2} \sum_{i\in S} (f(W,a,x_i)-y_i)^2,
$
where factor of $n$ in the objective function is included to normalize the sum of the losses across all samples.
\paragraph{Gradient Descent (GD).} We first demonstrate the standard GD optimization framework for training such network. Throughout the paper, for each $r\in[m]$, let $w_r(t)$ represent the weight vector $w_r$ at the $t$-th iteration. Then we have the update for $t+1$,
$
w_r(t+1) = w_r(t) - \eta\cdot\frac{ \partial L( W(t) ) }{ \partial w_r(t) }, r\in[m], 
$
where $\eta$ is the step size and $\frac{\partial L(W(t))}{\partial w_r(t)}$ has the following formulation 
\begin{align*}%\label{eq:gradient}
& ~ \frac{ \partial L(W(t)) }{ \partial w_r(t) } 
=  \frac{1}{ \sqrt{m} } \sum_{i=1}^n ( f(W(t), a, x_i) - y_i )  \cdot a_r \\
\cdot & ~ \mathbbm{1} ( w_r(t)^\top x_i >\tau  )\cdot x_i.
\end{align*}

\paragraph{Stochastic Gradient Descent (SGD).} In this work, we generalize the GD optimizer to SGD optimizer:
\begin{align}\label{eq:SGD_iter}
    w_r(t+1) = w_r(t) - \eta \cdot \frac{ \partial L(W(t),S_t) }{ \partial w_r(t) }, r\in[m], 
\end{align}
where the batch set $S_t$ is a uniform sub-sample of $[n]$. For simplicity, we define
\begin{align}\label{eq:def_noisy_G}
G_{t,r} := & ~ \frac{ \partial L(W(t), S_t) }{ \partial w_r(t)} \notag \\
= & ~ \frac{n}{|S_t|} \cdot \frac{1}{ \sqrt{m} } \sum_{i \in S_t} ( f(W(t), a, x_i) - y_i ) \notag \\
& ~ \cdot a_r \cdot\mathbbm{1} ( w_r(t)^\top x_i >\tau  )\cdot x_i.
\end{align}
\iffalse
Thus, we have
\begin{align}\label{eq:expected_sgd_is_gd}
    \E_{S_t \subset [n] } \Big[ \frac{ \partial L(W, S_t) }{ \partial w_r } \Big] =  \frac{ \partial L(W) }{ \partial w_r }
\end{align}
and with probability $1$ we have
\begin{align}\label{eq:absolute_sgd_bound}
     \|  \frac{ \partial L(W, S_t ) }{ \partial w_r }  \|_2 \leq \frac{n}{ |S_t| } \cdot \frac{1}{ \sqrt{m} } \cdot O(1) \cdot |S_t| \leq O( \frac{n}{\sqrt{m}} )
\end{align}
where the first step follows from triangle inequality and $|f(W,a,x_i)| = O(1)$ (by property of concentration of random Gaussian at the initialization) and $|y_i| = O(1)$ (by assumption of our problem). 
\Mingquan{Here it gives the bound of $\|G_{t,r}\|_2$ which will be renewed later.}
\fi
At iteration $t$, let 
%\begin{align}
$u(t):=[u_1(t), \cdots, u_n(t)]^\top\in\R^n$
%\end{align}
be the prediction vector, where each $u_i(t)$ satisfies that $u_i(t)=f(W(t),a,x_i)$, $i\in[n]$.    %%% Section 2. Problem formulation
\section{Asynchronous Tree and Kronecker Structured Data} \label{sec:technical_overview}
In this section, we will briefly describe the special data structure and input data assumption of the proposed algorithm. 

{\bf Asynchronous Tree Data Structure.} In each iteration of the training algorithm, there are two main parts: forward computation and backward computation. The goal of forward computation is to compute the prediction vector. By the property of the shift ReLU function, for each sampled data point $x_i$, we need to find which nodes in hidden layer are activated, which is a \textbf{query} operation. In backward computation, we need to compute the gradient vectors and then update the weight vectors, which is an \textbf{update} operation. 

In view of this situation, we propose the \textsc{AsynchronousTree} data structure which is a binary tree. It mainly supports \textbf{query} and \textbf{update} operations. Since the inner product of each weight vector $w_r$, $r\in[m]$ and input data point $x_i$, $i\in[n]$ is frequently compared with the threshold $\tau$, we maintain $n$ trees $T_1, \cdots, T_n$ for the $n$ data points respectively. For the $i$-th tree $T_i$, the values of leaf nodes of $T_i$ are the inner products of the $m$ weight vectors with $x_i$ and the value of each inner node is the maximum of the values of its two children. Hence, when executing the \textbf{query} operation in $T_i$, we start with the root of $T_i$ and recurse on its two children. Thus, given an index $i$ corresponding to point $x_i$ and a threshold $\tau$, a \textbf{query} operation finds all index $r\in[m]$ such that $w_r^\top x_i>\tau$. Let the number of activated nodes be $Q$, then the time for a batch of samples of \textbf{query} operation would be $O(Q\cdot\log{m})$ because the depth of each tree is $O(\log{m})$. Algorithm~\ref{alg:at_query_informal} shows \textbf{query} operation.

\begin{algorithm}[!ht]\caption{\textbf{Query} operations of asynchronous tree, informal version of Algorithm~\ref{alg:at_query}}\label{alg:at_query_informal}
\begin{algorithmic}[1]
\Procedure{Query}{$r=\mathrm{root}(T_i) \in T, \tau \in \R_{\geq 0}$} \Comment{\textsc{Query} in  Theorem~\ref{thm:correlation_tree_data_structure}} 
\If{$r$ is leaf}
\If{$r.\text{value}>\tau$}
\State \Return $r$
\EndIf
\Else
\State $r_1\gets$ left child of $r$, $r_2\gets$ right child of $r$
\If{$r_1.\text{value} > \tau$}
    \State $S_1 \gets $\textsc{Query}$(r_1, \tau)$
\EndIf
\If{$r_2.\text{value} > \tau$}
    \State $S_2 \gets $\textsc{Query}$(r_2, \tau)$
\EndIf
\EndIf
\State \Return $S_1 \cup S_2$
\EndProcedure
\end{algorithmic}
\end{algorithm}
% \fi
When some weight vector $w_r$, $r\in[m]$ changes, we need to recompute the inner products between $w_r$ and the $n$ data points and then update the $n$ corresponding trees, so the time for $n$ times \textbf{update} operations is $O(n\cdot(d+\log{m}))$. The Algorithm~\ref{alg:at_update} shows \textbf{update} operation.To be more specific, given a vector $z$ and index $r$, a \textbf{update} operation updates weight vector $w_r$ with $z$.

% \iffalse
\begin{algorithm}[!ht]\caption{\textbf{Update} operations of asynchronous tree, informal version of Algorithm~\ref{alg:at_update}}\label{alg:at_update_informal}
\begin{algorithmic}[1]
\Procedure{Update}{$z\in\R^d, r \in [m]$} \Comment{\textsc{Update} in  Theorem~\ref{thm:correlation_tree_data_structure}}
\For{$i=1$ to $n$} \label{lin:update_loop_informal}
    \State $l \gets$ the $r$-th leaf of tree $T_i$ \label{lin:update_find_leaf_informal}
    \State $l.\text{value} \gets l.\text{value}+z^\top x_i$ \label{lin:update_inner_product_informal}
    \While{$l$ is not root}
        \State $p$ $\gets$ parent of $l$
        \State $p.\text{value} \gets \max \{ p.\text{value}, l.\text{value} \}$
        \State $l \gets p$
    \EndWhile
\EndFor
\EndProcedure
\end{algorithmic}
\end{algorithm}
% \fi

Note that the above statements are for the general case, that is, there are no requirements for the input data. Since we use the stochastic gradient descent method, each iteration randomly selects a batch $S_t$ from the set $[n]$. The \textbf{query} operation is executed for the trees whose indexes are in the set $S_t$ but not all the $n$ trees, that is why this data structure is called \textsc{AsynchronousTree}. We give a formal statement discussing \textsc{AsynchronousTree}'s operations' time cost in Theorem~\ref{thm:correlation_tree_data_structure}, Section~\ref{sec:asynchronize_tree} 

\iffalse
\paragraph{Asynchronous Tree Data Structure.} The initial algorithm stated above is quite straightforward. We notice that there are $B$ data points are sampled because of SGD algorithm, so when some $w_r$, $r\in[m]$ are changed, we do not need to update all the $n$ trees. For each data point $x_i$, $i\in[n]$, let $c_i\in\R$ be the time stample, which records the time that the $i$-th data point is sampled. We construct a tree $\mathcal{T}$ such that the $m$ leaf nodes represent the time of the $m$ weight vectors has changed. Then for each sampled point $x_i$, we first query the set $R_{t,i}$ of weight vectors that have changed but have not been updated in $T_i$. Next for each $r\in R_{t,i}$, we update the tree $T_i$ based on each weight vector $w_r$. Then in the forward computation, let $x_i$ call the query procedure to find the set of neurons that are activated and compute the prediction value. The convenience for us is that in the backward computation, we do not need to update all the $n$ trees, and only to update the tree $\mathcal{T}$ since some weight vectors are changed. Since the $n$ trees are not changed simultaneously, we call this data structure \textsc{AsynchronousTree}. 
\fi

{\bf Kronecker Structured Data.} Recall that in the \textbf{update} operation of data structure \textsc{AsynchronousTree}, we need to compute the inner products between $w_r$ and the $n$ data points, i.e., the quantity $X^\top w_r$, where $X=[x_1, \cdots, x_n]$. When the data points have Kronecker property such that $x_i=\vect(\ov{x}_i\ov{x}_i^\top)\in\R^d$ for each $i\in[n]$, it would be efficient to compute the matrix-vector multiplication $X^{\top}w_r$. In particular, we have the following equation
\begin{align*}
     (X^\top w_r )_i= (\ov{X}^\top\cdot\vect^{-1}(w_r)\cdot\ov{X} )_{i,i},
\end{align*}
where $\ov{X}=[\ov{x}_1, \cdots, \ov{x}_n]$ and $\vect^{-1}(\cdot)$ is the inverse vectorization operator. Then the computing of $X^\top w_r$ is transferred to the fast matrix multiplication. 

At the $t$-th iteration, we need to compute the gradient vector denoted as $\delta_{t,r}$ to update the weight vector $w_r$ with $r\in[m]$, that is, $w_r(t+1)=w_r(t)+\delta_{t,r}$. When $w_r$ changes, we need to recompute $w_r^\top x_i$ for $i\in[n]$. Since $w_r(t)^\top x_i$ is already known, in order to compute $w_r(t+1)^\top x_i$, we only need to compute $\delta_{t,r}^\top x_i$. To be specific, the vector $\delta_{t,r}$ has such form $\delta_{t,r}=X_{:, S_t}\cdot c$ with $c\in\R^{|S_t|}$, then we have $\delta_{t,r}^\top x_i=c^\top X_{:,S_t}^\top x_i$, where for $j\in[|S_t|]$,
\begin{align*}
   (X_{:,S_t}^\top x_i )_j= (\ov{X}_{:,S_t}^\top\cdot(\ov{x}_i\ov{x}_i^\top)\cdot\ov{X}_{:,S_t} )_{j,j}. 
\end{align*}

Hence, the computation of vector $X_{:,S_t}^\top x_i$ is reduced to the pairwise inner products $\ov{x}_i^\top\ov{x}_j$ for $i,j\in[n]$, which can be precomputed at the initialization, and takes time only $O(|S_t|)$. Now the \textbf{update} operation can be completed by computing $\delta_{t,r}^\top x_i$ for all $i\in[n]$ and then updating the $n$ trees, and thus takes time $O(n\cdot(|S_t|+\log{m}))$, which is faster than the fast matrix multiplication.

\iffalse
Recall that in the data structure \textsc{SearchTree}, the update operation computes the inner products between the $n$ data points and a weight vector $w_r$, which is equivalent to computing $X^\top w_r$, i.e., the diagonal of the matrix $\ov{X}^\top\cdot\vect^{-1}(w_r)\cdot\ov{X}$. Correspondingly, the query operation in \textsc{SearchTree} can be improved from $O(n\cdot(d+\log{m}))$ to
\begin{align*}
O(n\cdot(d^{\frac{\omega-1}{2}}+\log{m})),
\end{align*}
where $\omega$ is the exponent of matrix multiplication. In the end, each iteration in the training algorithm can be completed in time $n\cdot o(m)\cdot o(d)$. 
\fi

\vspace{-1mm}
\section{Efficient SGD Algorithm}\label{sec:efficient_sgd}

In this section, we present our training algorithm. The algorithm for training a two-layer fully connected neural network using SGD and Asynchronous tree data structure is shown in Algorithm~\ref{alg:sgd_at} which has two main parts: the \textit{initialization}(see Section~\ref{sec:asynchronize_tree}) step and the \textit{for} loop of iterations. 

\iffalse
The initialization step initializes the $m$ weight vectors $w_1, \cdots, w_m\in\R^d$ and computes the inner products $w_r^\top x_i$ for $r\in[m]$ and $i\in[n]$. At each iteration $t\ge 0$, the \textit{forward} step (Line~\ref{line:S_t}-\ref{line:end_forward}) computes the prediction vector $u(t)\in\R^n$; given the sampled set $S_t\subset [n]$, for some $x_i$ with $i\in S_t$, we need to look up the activated neurons such that  $w_r^\top x_i>\tau$, which is implemented by the \textsc{Query} procedure of the asynchronous tree data structure; in addition, the set of activated neurons $L_i$ has the size bound shown in Lemma~\ref{lem:k_it}. The \textit{backward} step (Line~\ref{line:ell_t}-\ref{line:end_backward}) updates the weight vectors; the set of weight vectors that would be changed is denoted as $\ell(t)$, whose size has a relationship with the the size of $L_i$ (see Lemma~\ref{lem:K=g(Q)}); the incremental vector for weight vector $w_r$ is denoted as $\delta_{t,r}$; since $w_r$ changes, we need to update all the inner products $w_r^\top x_i$ for $i\in[n]$, which is executed by the \textsc{Update} procedure of the asynchronous tree data structure.  
\fi

\begin{algorithm*}[!ht] 
\caption{Accelerate computation in each iteration using asynchronous tree data structure}
\label{alg:sgd_at}
\begin{algorithmic}[1]
	\Procedure{OurAlgorithm}{$X=[x_1, \cdots, x_n]\in\R^{d\times n}$, $y\in\R^n$}
	\State Initialize $w_r(0)\sim\N(0, I_d)$ for each  $r\in[m]$
	\State Construct a \textsc{AsynchronousTree} data structure \textsc{AT}
	\State \textsc{AT}.\textsc{Init}($\{w_r(0)\}_{r\in[m]}$, $\{x_i\}_{i\in[n]}$, $n$, $m$, $d$)
	\For{$t=1$ to $T$}
	    %\State {\color{blue}/*Forward Computation*/}
	    \State Sample a set $S_t \subset [n]$ with $|S_t|=S_{\batch}$ uniformly at random \label{line:S_t}
	    \For{each $i \in S_t $}
	    \State $L_i\gets\textsc{AT}.\textsc{Query}(i, \tau)$\Comment{$|L_i|\le Q$} \label{line:st_Li_formal}
	    \State $u_i(t)\gets\frac{1}{\sqrt{m}}\sum_{j\in L_i}a_j\cdot\phi_\tau(j.\text{value})$\Comment{$u_i(t)$ is the $i$-th entry of vector $u(t)$}\label{line:st_uit_formal}
	    \EndFor\label{line:end_forward}
	    %\State {\color{blue}/*Backward Computation*/}
		\State $\ell(t) \gets \cup_{i\in S_t} L_i$\Comment{$\ell(t)\subset [m]$ is the index set such that $w_r$ changes for each $r\in\ell(t)$ and $|\ell(t)|\le K$}\label{line:ell_t}
		\For{each $r\in\ell(t)$}
			\State $v\gets 0_n$ and $v_i\gets\frac{\eta n}{S_{\batch}\sqrt{m}}\cdot a_r\cdot\mathbbm{1}(w_r(t)^\top x_i>\tau)$ for $i\in S_t$ 
                \State \Comment{The factor of $n$ is to normalize the sum of the losses across all samples}
			\State $\delta_{t,r}\gets X_{:, S_t}\cdot(v_{S_t}\odot(y-u(t))_{S_t})$\Comment{$\delta_{t,r}=-\eta\cdot G_{t,r}$}\label{line:st_delta_tr_formal}
			\State \textsc{AT}.\textsc{Update}($\delta_{t,r}$, $r$)\label{line:st_update_formal}
		\EndFor \label{line:end_backward}
	\EndFor
	\State \Return $u(T)$
	\EndProcedure
\end{algorithmic}
\end{algorithm*}	

In our algorithm, the initialization step initializes the $m$ weight vectors $w_1, \cdots, w_m\in\R^d$ and computes the inner products $w_r^\top x_i$ for $r\in[m]$ and $i\in[n]$. At each iteration $t\ge 0$, the \textit{forward} step (Line~\ref{line:S_t}-\ref{line:end_forward}) computes the prediction vector $u(t)\in\R^n$; given the sampled set $S_t\subset [n]$, for some $x_i$ with $i\in S_t$, we need to look up the activated neurons such that  $w_r^\top x_i>\tau$, which is implemented by the \textsc{Query} procedure of the asynchronous tree data structure; in addition, the set of activated neurons $L_i$ has the size bound shown in Lemma~\ref{lem:k_it}. The \textit{backward} step (Line~\ref{line:ell_t}-\ref{line:end_backward}) updates the weight vectors; the set of weight vectors that would be changed is denoted as $\ell(t)$, whose size has a relationship with the the size of $L_i$ (see Lemma~\ref{lem:K=g(Q)}); the incremental vector for weight vector $w_r$ is denoted as $\delta_{t,r}$; since $w_r$ changes, we need to update all the inner products $w_r^\top x_i$ for $i\in[n]$, which is executed by the \textsc{Update} procedure of the asynchronous tree data structure.

\section{Convergence and Complexity Analysis}\label{sec:convergence_complextiy}

So far, we have presented how to efficiently train the two-layer fully connected neural network by SGD. In this section, we present the convergence of the algorithm, the lower bound of the width of neural network and analyze the cost e.g., the complexity, for each iteration. In Section~\ref{sec:data_with_kronecker:convergence}, we give the convergence theorem of Algorithm~\ref{alg:sgd_at}. We introduce some useful properties when the input data points have Kronecker structure in Section~\ref{sec:data_with_kronecker:preliminary} and then prove the formal version of Theorem~\ref{thm:main_informal} and give its general case in Section~\ref{sec:data_with_kronecker:Kronecker_formal}.

\subsection{Convergence Analysis}\label{sec:data_with_kronecker:convergence}

We start with the definition of Gram matrix, which can be found in \cite{dzps19}.
\begin{definition}[Data-dependent matrix $H$]\label{def:data_dependent_function}
Given a collection of data points $\{x_1, \cdots, x_n \} \subset \R^d$ and $m$ weight vectors $\{w_1, \cdots, w_m\}\subset\R^d$, the continuous (resp. discrete) Gram matrix denoted as $H^{\cts}$ (resp. $H^{\dis}$) is defined by 
% \begin{align*}
$
    H_{i,j}^{\cts}:=\E_{w\sim\N(0, I_d)} [x_i^\top x_j\cdot \mathbbm{1}( w^\top x_i > \tau, w^\top x_j >\tau ) ]
$,
$
    H_{i,j}^{\dis}:=\frac{1}{m}\sum_{r=1}^{m}x_i^\top x_j\cdot\mathbbm{1} (w_r^\top x_i>\tau, w_r^\top x_j>\tau ).
$
% \end{align*}
Let $\lambda:=\lambda_{\min}(H^{\cts})$ be the smallest eigenvalue of the matrix $H^{\cts}$ and assume $\lambda\in(0,1]$. 
\end{definition}

\begin{remark}
For more detailed discussion about the assumption of $\lambda$, we refer the readers to~\cite{dzps19}. This assumption is commonly used in \cite{sy19,dllwz19,syz21,szz24,bpsw21,hlsy21,als+23,szz22,gms23,yjz+23}.
\end{remark}
 
Given the two matrices $H^{\cts}$ and $H^{\dis}$, the following lemma gives the bound of $\lambda_{\min}(H^{\dis})$. Previous work implies the following standard result \cite{dzps19,syz21}.
\begin{lemma}[Lemma C.1 in~\cite{syz21}] \label{lem:H_dis_lambda_min}
For any shift threshold $\tau\geq 0$, let $\lambda := \lambda_{\min}(H^{\cts})$ and $m=\Omega (\lambda^{-1}n\log(n/\alpha) )$ be the number of samples in $H^{\dis}$, then 
$
\Pr [\lambda_{\min}(H^{\dis})\ge\frac{3}{4}\lambda ]\ge 1-\alpha.
$
\end{lemma}
The Lemma is also valid in our settings. Besides the two matrices $H^{\cts}$ and $H^{\dis}$, each iteration $t\ge 0$ has a data-dependent matrix $H(t)$ defined below. 
\begin{definition}[Dynamic data-dependent matrix $H(t)$]
For $t\ge 0$, given the $m$ weight vectors $\{w_1(t), \cdots, w_m(t)\}\subset\R^d$ at iteration $t$, the corresponding data-dependent matrix $H(t)$ is defined by 
% \begin{align*}
$
    H(t)_{i,j}:=\frac{1}{m}\sum_{r=1}^{m}x_i^\top x_j\cdot\mathbbm{1} (w_r(t)^\top x_i>\tau, w_r(t)^\top x_j>\tau).
$
% \end{align*}
\end{definition}

Finally, we present the convergence theorem for our algorithm.
%The algorithm for training a two-layer fully connected neural network using SGD is shown in Algorithm~\ref{alg:ds_sgd} shown in Appendix~\ref{sec:ds_sgd_alg}, where the \textsc{CorrelationTree} data structure is placed in Appendix~\ref{sec:correlation_tree}. For Algorithm~\ref{alg:ds_sgd}, we have the following theorem. \Mingquan{Note the correspondence relationship between 9-page formal version and the appendix.}
\begin{theorem}\label{thm:quartic}
Given $n$ training samples $\{(x_i, y_i)\}_{i=1}^{n}$ and a parameter $\rho\in(0,1)$. Initialize $w_r\sim\N(0, I_d)$ and sample $a_r$ from $\{-1,+1\}$ uniformly at random for each $r\in[m]$. Set the width of neural network to be %\Zhao{Write below as poly. No need to write exponent.} 
$
    m=\poly (\lambda^{-1}, S_{\batch}^{-1}, n, \log(n/\rho) ),
    %m=\Omega(\lambda^{-4}B^{-2}n^6 \log^3(n/\rho))
$
and the step size
$
\eta=\poly(\lambda, S_{\batch}, n^{-1}),
$
where $\lambda=\lambda_{\min}(H^{\cts})$ and $S_{\batch}$ is the batch size, then with probability at least $1-O(\rho)$, the vector $u(t)$ for $t\ge 0$ in Algorithm~\ref{alg:sgd_at} satisfies that 
\begin{align}\label{eq:quartic_condition}
  \E [\| u (t) - y \|_2^2 ]  \leq ( 1 - \eta \lambda / 2 )^t \cdot \| u (0) - y \|_2^2 .
\end{align}
\end{theorem}
The proof for this theorem is deferred to Section~\ref{app:proof_convergence}. 

\subsection{Preliminary for Complexity Analysis}\label{sec:data_with_kronecker:preliminary}
After proving the convergence of our algorithm, we analyze the running time cost. Before giving our main theorem, we need some preliminary information. The set of neurons that are activated (i.e., $w_r^\top x_i>\tau$) in Algorithm~\ref{alg:sgd_at} denoted as $L_i$ is formally defined in the following definition.  
\begin{definition}[Fire set]
For each $i\in[n]$ and $0\le t\le T$, let $\mathcal{S}_{i, \fire}\subset[m]$ denote the set of neurons that are ``fired" at time $t$, i.e., 
$
\mathcal{S}_{i,\fire}(t):= \{r\in[m]\mid w_r(t)^\top x_i>\tau \}. 
$
\end{definition}
Let $k_{i,t}:=|\mathcal{S}_{i, \fire}(t)|$, then the following lemma gives the upper bound of $k_{i,t}$. 
\begin{lemma}[Lemma C.10 in \cite{syz21}]\label{lem:k_it}
For $0<t\le T$, with probability at least 
$
1-n\cdot\exp(-\Omega(m)\cdot\min\{R, \exp(-\tau^2/2)\}), 
$
we have the following holds for all $i \in [n]$
$
k_{i,t}=O(m\cdot\exp(-\tau^2/2))
$
where $R$ is a parameter that depends on $m$, $n$, and $\lambda$. 
\end{lemma}

The Lemma is also valid in our settings. By Lemma~\ref{lem:k_it}, in our setting, we have that with high probability, $Q=O(m\cdot\exp(-\tau^2/2))$. Moreover, the set of changed weight vectors denoted as $\ell(t)$ in Algorithm~\ref{alg:sgd_at} satisfies that the upper bound of its size $K$ has the following relationship with the quantity $Q$. 
\begin{lemma}\label{lem:K=g(Q)}
The parameters $K$ and $Q$ in Algorithm~\ref{alg:sgd_at} satisfy that $K=O(S_{\batch}\cdot Q)$. 
\end{lemma}

\begin{proof}
In Algorithm~\ref{alg:sgd_at}, the weight vector $w_r$ is updated if $G_{t,r}\neq 0$, then there exists at least one $i\in S_t$ such that $w_r(t)^{\top}x_i>\tau$. Since for each $i\in[n]$, there are at most $Q$ neurons that are activated; in addition, $|S_t|=S_{\batch}$, thus there are at most $S_{\batch}\cdot Q$ weight vectors that are changed. 
\end{proof}
\paragraph{Properties of Kronecker Structure and Related Computation Facts}\label{sec:Kronecker}
Before proving the formal version of Theorem~\ref{thm:main_informal}, we provide some background about matrix multiplication and Kronecker product. Let the time of multiplying two matrices in $\R^{a \times b}$ and $\R^{b \times c}$ be $\Tmat(a,b,c)$. In particular, we use $\omega$ to denote the exponent of matrix multiplication, which means that $ \Tmat(n,n,n) = n^\omega$. Currently, $\omega \approx 2.373$~\cite{williams12,le14}. For the time of matrix multiplication $\Tmat(a, b, c)$, we have the two following properties. 
\begin{fact}[\cite{bcs97,b13}]\label{fact:Tmat_permutation}
$\Tmat(a, b, c) = O( \Tmat(a, c, b) ) =O(\Tmat(c, a, b))$. %=\Tmat(b, c, a)=\Tmat(c, a, b)=\Tmat(c, b, a)$. 
\end{fact}
The above has been widely used in optimization and dynamic algorithms, e.g., see \cite{cls19,lsz19,bns19,jkl+20,b20,sy21,hjstz21,dsw22,gs22,dsw22,sswz22_lattice,syyz22,sswz23_quartic,as23,dms23,bsz24,dls23,lsz23,qszz23,gsy25,syyz23} as an example.

\begin{fact}\label{fact:Tmat_monotonicity}
For any $c \geq d>0$, $\Tmat(a, b, c) \geq \Tmat(a, b, d)$. 
\end{fact}

Then, we state tensor tools utilized in our work.
\begin{claim}[Tensor trick]\label{cla:tensor_trick}
Given a matrix $H\in\R^{d\times d}$, let $h:=\vect(H)\in\R^{d^2}$. Given a matrix $V\in\R^{d\times n}$, the matrix $U\in\R^{n\times d^2}$ is defined satisfying that the $i$-th row of $U$ is equal to $ (\vect(v_i v_i^\top) )^\top$, where $v_i\in\R^d$ is the $i$-th column of $V$. Then for each $i\in[n]$, it holds that  
%\begin{align*}
     $(V^\top H V )_{i,i} = ( U \cdot h )_{i}.$
%\end{align*}
\end{claim}

The proof is in Section~\ref{sec:notation_tools}. It tells us that given $U\in\R^{n\times d^2}$ and $h\in\R^{d^2}$, the computation of $(Uh)_i$ with $i\in[n]$ has an equivalent way when each row of $U$ has the form $U_{i,:}^\top=\vect(xx^\top)$ for some $x\in\R^d$. 

By virtue of the property presented in Claim~\ref{cla:tensor_trick}, given $L\subseteq [n]$, we can compute $(Uh)_L$ efficiently using the following lemma. 
\begin{lemma}[Improved running time via tensor trick]\label{lem:tensor_trick_rt}
Let $U$ and $h$ be defined same as in Claim~\ref{cla:tensor_trick}, given $L\subseteq [n]$, then computing $(U h)_L$ takes time 
\begin{align*}
    \begin{cases}
    \Tmat(d,|L|,d), & \mathrm{if~} |L| \leq d;\\
    (|L|/d) \cdot \Tmat(d,d,d), & \mathrm{otherwise.} 
    \end{cases}
\end{align*}
\end{lemma}
\begin{proof}
For the case $|L|=d$, it can be computed in $\Tmat(d,d,d) = d^\omega$ time. To see that, without of loss generality, assume $L = [d]$. By Claim~\ref{cla:tensor_trick}, for each $i\in[d]$, we have $(Uh)_i=(V^\top H V)_{i,i}$. Therefore, the computation of $(Uh)_L$ is reduced to computing 
$V^\top H V$ first and then taking the diagonal entries, which takes time $d^\omega$ that is faster than $d^3$. In a similar way, 
For $|L| < d$, we can compute it in $\Tmat(|L|, d, d)+\Tmat(|L|, d, |L|)$ time, which is equal to $\Tmat(d,|L|,d)$ by Fact~\ref{fact:Tmat_permutation} and Fact~\ref{fact:Tmat_monotonicity}. For $|L|>d$, we can divide $L$ into $|L|/d$ groups and each one is reduced to $|L|=d$ case. Thus it can be computed in $|L|/d \cdot \Tmat(d,d,d)$ time.
\end{proof}

\subsection{Complexity Theorem}\label{sec:data_with_kronecker:Kronecker_formal}

So far, we have presented necessary preliminary information. Finally, we could give our main theorem, which shows the time cost of Algorithm~\ref{alg:sgd_at} . 
\begin{theorem}[Formal version of Theorem~\ref{thm:main_informal}]\label{thm:main_formal}
Given $n$ training samples $\{(x_i, y_i)\}_{i=1}^{n}$ such that $x_i=\vect(\ov{x}_i\ov{x}_i^\top)\in\R^d$ and $y_i\in\R$ for each $i\in[n]$,let $\omega $ denote matrix multiplication exponent, there exists a SGD algorithm that can train a two-layer fully connected neural network such that the initialization takes time  
$
    O(m\cdot n\cdot d^{\frac{\omega-1}{2}})
$
and with high probability each iteration takes time 
$
    S_{\batch}^2\cdot o(m)\cdot n, 
$
where $S_{\batch}$ is the batch size, $m$ is the width of the neural network, and $\omega$ is the exponent of matrix multiplication.
\end{theorem}

\begin{proof}
Let $\ov{X}:=[\ov{x}_1, \cdots, \ov{x}_n]\in\R^{\sqrt{d}\times n}$ and
$X:= [\vect(\ov{x}_1\ov{x}_1^\top), \cdots, \vect(\ov{x}_n\ov{x}_n^\top) ]\in\R^{d\times n}$. In the initialization step, there are two main parts: $(1)$ in order to construct the $n$ trees $T_i$, $i\in[n]$, we need to compute $X^\top w_r$ for all $r\in[m]$; $(2)$ the pairwise inner products $\ov{x}_i^\top\ov{x}_j$ for all $i,j\in[n]$, which will be used in the backward computation. 

By Claim~\ref{cla:tensor_trick}, given a fixed $r\in[m]$, we have that 
\begin{align*}
     (X^\top w_r )_i= (\ov{X}^\top\cdot\vect^{-1}(w_r)\cdot\ov{X} )_{i,i},\ i\in[n]. 
\end{align*}
By Lemma~\ref{lem:tensor_trick_rt}, we can compute the matrix $\ov{X}^\top\cdot\vect^{-1}(w_r)\cdot\ov{X}$ and then take its diagonal.

Assume that $n>\sqrt{d}$, then we can compute $X^\top w_r$ in time $(n/\sqrt{d} )\cdot\Tmat (\sqrt{d}, \sqrt{d}, \sqrt{d} )=O (n\cdot d^{\frac{\omega-1}{2}} )$. Hence the first part takes time $O (m\cdot n\cdot d^{\frac{\omega-1}{2}} )$. In addition, the second part takes time $O (n^2\cdot\sqrt{d} )$. Since $m=\poly(n)$, the initialization step takes time $O (m\cdot n\cdot d^{\frac{\omega-1}{2}} )$. 

Now we analyze the time complexity of each iteration in Algorithm~\ref{alg:sgd_at}. 

\textbf{Forward Computation.} Line~\ref{line:st_Li_formal} takes time $O(Q\cdot\log{m})$. Line~\ref{line:st_uit_formal} takes time $O(Q)$. Hence the forward computation takes time $O(S_{\batch}\cdot Q\cdot\log{m})$. 

\textbf{Backward Computation.} In Line~\ref{line:st_delta_tr_formal}, the computation of $v_{S_t}$ and $(y-u(t))_{S_t}$ takes time $O(S_{\batch})$. In Line~\ref{line:st_update_formal}, we need to compute $\delta_{t,r}^\top x_i$ for each $i\in[n]$, in which the core part is to compute the product $X_{:, S_t}^\top x_i$. By Claim~\ref{cla:tensor_trick}, we have that for each $j\in S_t$, 
$
     (X_{:, S_t}^\top x_i )_j= (\ov{X}_{:,S_t}^\top\cdot(\ov{x}_i\ov{x}_i^\top)\cdot\ov{X}_{:,S_t} )_{j,j}, 
$
which only needs the pairwise products $x_i^\top x_j$, $i,j\in[n]$ that are already computed in initialization step. Hence, Line~\ref{line:st_update_formal} takes time $O(n\cdot(S_{\batch}+\log{m}))$ and the backward computation takes time $O(K\cdot n\cdot (S_{\batch}+\log{m}))$. 

By Lemma~\ref{lem:K=g(Q)} and setting $\tau=\sqrt{(\log{m})/2}$, we have that each iteration takes time  
\begin{align*}
  & ~ O (S_{\batch}^2\cdot m^{3/4}\cdot n+S_{\batch}\cdot m^{3/4}\cdot n\cdot\log{m} ) \\
= & ~ S_{\batch}^2\cdot o(m)\cdot n, 
\end{align*}
% \begin{align*}
% O (S_{\batch}^2\cdot m^{3/4}\cdot n+S_{\batch}\cdot m^{3/4}\cdot n\cdot\log{m} ) = S_{\batch}^2\cdot o(m)\cdot n,
% \end{align*}
which is independent of the data dimensionality $d$. 
\end{proof}

Moreover, we have the following corollary which is a general version of Theorem~\ref{thm:main_formal}. 
\begin{corollary}\label{cor:general_tensor}
Given $n$ training samples $\{(x_i, y_i)\}_{i=1}^{n}$ such that for each $i\in[n]$, $x_i=b_i\otimes a_i\in\R^d$ and $a_i\in\R^p$, $b_i\in\R^q$, and $p,q=O(\sqrt{d})$, let $\omega $ denote matrix multiplication exponent, there exists an SGD algorithm that can train a two-layer fully connected neural network such that the initialization takes time 
$
O (m\cdot n\cdot d^{\frac{\omega-1}{2}} )
$
and with high probability each iteration takes time $
S_{\batch}^2\cdot o(m)\cdot n,
$
where $S_{\batch}$ is the batch size, $m$ is the width of the neural network, and $\omega$ is the exponent of matrix multiplication.
\end{corollary}

%The proof of Corollary~\ref{cor:general_tensor} is in Section ~\ref{sec:cor_proof}.

\begin{proof}
Let $A:=[a_1, \cdots, a_n]\in\R^{p\times n}$, $B:=[b_1, \cdots, b_n]\in\R^{q\times n}$, and $X:=[x_1, \cdots, x_n]\in\R^{d\times n}$. Similar to Theorem~\ref{thm:main_formal}, in the initialization step, there are two main parts: $(1)$ computing $X^{\top}w_r$ for all $r\in[m]$ to construct the $n$ trees $T_i$, $i\in[n]$; $(2)$ computing the pairwise inner products $a_i^\top a_j$, $b_i^\top b_j$ for all $i,j\in[n]$ that will be used in the backward computation. 

By Claim~\ref{cla:general_tensor}, for a fixed $r\in[m]$, it holds that 
$
     (X^\top w_r )_i= (A^\top\cdot\vect^{-1}(w_r)\cdot B )_{i,i},\ i\in[n].
$
By Lemma~\ref{lem:tensor_trick_rt}, we can compute the matrix $A^\top\cdot\vect^{-1}(w_r)\cdot B$ and then take its diagonal. Assume that $n=O(\sqrt{d})$, then we can compute $X^\top w_r$ in time
\begin{align*}
O (n/\sqrt{d} )\cdot\Tmat (\sqrt{d}, \sqrt{d}, \sqrt{d} )=O (n\cdot d^{\frac{\omega-1}{2}} )
\end{align*}

since $p,q=O(\sqrt{d})$. Hence the first part takes time $O (m\cdot n\cdot d^{\frac{\omega-1}{2}} )$. Furthermore, the second part takes time $O (n^2\cdot\sqrt{d} )$. Since $m=\poly(n)$, the initialization step takes time $O (m\cdot n\cdot d^{\frac{\omega-1}{2}} )$. 

Now we analyze the time complexity of each iteration in Algorithm~\ref{alg:sgd_at}.

\textbf{Forward Computation.} It is same as the forward computation in Theorem~\ref{thm:main_formal}, i.e., $O(S_{\batch}\cdot Q\cdot\log{m})$.

\textbf{Backward Computation.} In Line~\ref{line:st_delta_tr_formal}, the computation of $v_{S_t}$ and $(y-u(t))_{S_t}$ takes time $O(S_{\batch})$. In Line~\ref{line:st_update_formal}, we need to compute $\delta_{t,r}^\top x_i$ for each $i\in[n]$, in which the core part is to compute the product $X_{:, S_t}^\top x_i$. By Claim~\ref{cla:general_tensor}, we have that for each $j\in S_t$, 
\begin{align*}
     (X_{:, S_t}^\top x_i )_j 
     = & ~ (A_{:, S_t}^\top\cdot\vect^{-1}(x_i)\cdot B_{:, S_t} )_{j,j} \\
     = & ~ (A_{:, S_t}^\top\cdot a_i\cdot b_i^\top\cdot B_{:, S_t} )_{j,j},
\end{align*}
where the second step follows by $x_i=\vect(a_i b_i^\top)$. Hence we only need to compute the pairwise products $a_i^\top a_j$, $b_i^\top b_j$ for $i,j\in[n]$ which are already computed in initialization step. Then Line~\ref{line:st_update_formal} takes time $O(n\cdot(S_{\batch}+\log{m}))$ and the backward computation takes time 
$
O(K\cdot n\cdot(S_{\batch}+\log{m})).
$
Same as Theorem~\ref{thm:main_formal}, each iteration takes time 
$
S_{\batch}^2\cdot o(m)\cdot n, 
$
which is independent of the data dimensionality $d$. 
\end{proof}
%\vspace{-4mm}
\section{Discussion and Conclusion}\label{sec:discussion}
In this work, we propose the asynchronous SGD algorithm for training a two-layer fully connected neural network. As a lot of existing work about over-parameterized neural networks, we also prove the convergence of our training algorithm. In addition, in contrast with the existing work which improve training cost in each iteration by taking advantage of LSH technique or some data structures on space partitioning, we consider accelerating the computations in each iteration from the point of input data. When the input data have some special structures, e.g., Kronecker property, we propose the asynchronous tree data structure that completes the computations in each iteration in $o(m)\cdot n$ time, which is independent of the data dimension $d$.

With the Kronecker structure, one could utilize the form of the gradient to write the update as learning an $|S_t|$-dimensional coefficients for the diagonal of a covariance matrix of the data, which could be precomputed. Hence, one only needs to read $|S_t|$ entries from the precomputed covariance matrix then linearly combine them with $|S_t|$ coefficients. In contrary, without the Kronecker structure, we no longer have the nice decomposition of gradient updates into a linear combination of diagonals of a covariance matrix, hence we would have to spend $d$ time to recompute the inner product, hence an update time of $O(n (d+\log m))$.

Moreover, we would like to emphasize the importance of getting rid of the factor $d$ in the update time with the Kronecker structure. Note that since we assume the input is from a product space where $d=d_1\times d_2$, $d$ could be prohibitively large. Even for $d_1=d_2=384$ where there are standard choices of dimensions one encounters in practice, $d=384^2\approx 150,000$ which is much larger than individual dimensions. Hence, it’s crucial to design an algorithm whose update time does not depend on $d$ at all. On the other hand, $|S_t|$ is the batch size, which could be as small as 1 (for SGD). Thus, in most scenarios, one would prefer an algorithm that depends on $|S_t|$ instead of $d$.

The dependence on $d$ in turn appears at the initialization phase, where we spend $O(mnd^{(\omega-1)/2})$ time to initial the data structure. We would like to examine this runtime under two settings: 1).\ in practice where $\omega=3$, our initialization time is $O(mnd)$, we note that this is the size of input data if data is given to us in their Kronecker product form, or the per-iteration runtime of prior works \cite{dzps19}. In contrast, we only need to pay this runtime once instead for each iteration. 2).\ in theory, the common belief is that $\omega=2$, in this setting, our initialization time is $O(mnd^{1/2})$, this is precisely the size of input data if we assume $d_1=d_2=d^{1/2}$. A simple information theoretical argument would reveal that one has to spend $mnd^{1/2}$ to read the input data. Hence, in this setting, our initialization time is nearly optimal.

\section*{Impact Statement}

This paper presents work whose goal is to advance the field of Machine
Learning. There are many potential societal consequences of our work, none
which we feel must be specifically highlighted here.

\ifdefined\isarxiv
\bibliographystyle{alpha}
\bibliography{ref}
\else
\bibliographystyle{icml2026}
\bibliography{ref}
\fi

%%% The part below is the appendix

\newpage
\onecolumn
\appendix

\begin{center}
    \textbf{\LARGE Appendix }
\end{center}

% \section*{Appendix} %%%Zhao: don't delete this, this is required. Junwei: Comment this for AISTATS 2025

\paragraph{Roadmap.} We present the notations and tools used throughout the appendix~\ref{sec:notation_tools}. The missing proofs for some technical claims are shown in Section~\ref{sec:missing_proof}. In Section~\ref{app:proof_convergence}, we complete the proof of convergence theorem (Theorem~\ref{thm:quartic}). Next, the asynchronous tree data structure is presented in Section~\ref{sec:asynchronize_tree}. 
% Finally, we discuss the limitation of this work in Section~\ref{sec:limitation} and potential impacts in Section~\ref{sec:impact}.

\section{Notations and Tools}\label{sec:notation_tools}
% The section organized as follows: In Section~\ref{sec:notations}, we introduce notations that are used frequently in the appendix. 
Then, we introduce concentration inequality tools in Section~\ref{sec:prob_tool}. Then we present the missing proof of Claim~\ref{cla:tensor_trick} in Section~\ref{sec:notation_tools:tensor_tool} and general cases tensor tools which are used in Corollary~\ref{cor:general_tensor} in Section~\ref{sec:notation_tools_general_tensor_tools}.

% \subsection{Notations}\label{sec:notations}

% For a positive integer $n$, let $[n]$ represent the set $\{1, 2, \cdots, n\}$. For a matrix $A\in\R^{d\times n}$ and a subset $S\subset [n]$, $A_{i,j}$ is the entry of $A$ at the $i$-th row and the $j$-th column, and $A_{:, S}$ represents the matrix whose columns correspond to the columns of $A$ indexed by the set $S$. Similarly, for a vector $x\in\R^n$, $x_{S}$ is a vector whose entries correspond to the entries in $x$ indexed by the set $S$. Let $\|\cdot\|_2$ and $\|\cdot\|_F$ represent the $\ell_2$ norm and Frobenius norm respectively. The symbol $\mathbbm{1}(\cdot)$ represents the indicator function. For a positive integer $d$, $I_d$ denotes the $d\times d$ identity matrix. For a random variable $X$, let $\E[X]$ denote the expectation of $X$. The symbol $\Pr[\cdot]$ represents probability. 

\subsection{Probability Tool}\label{sec:prob_tool}

We state a general concentration inequality tool, which can be viewed as a more general version of Chernoff bound \cite{c52} and Hoeffding bound \cite{h63}.
\begin{lemma}[Bernstein inequality \cite{b24}]\label{lem:bernstein}
Let $X_1, \cdots, X_n$ be independent zero-mean random variables. Suppose that $|X_i| \leq M$ almost surely, for all $i$. Then, for all positive $t$,
\begin{align*}
\Pr [ \sum_{i=1}^n X_i > t] \leq \exp( - \frac{ t^2/2 }{ \sum_{j=1}^n \E[X_j^2]  + M t /3 }).
\end{align*}
\end{lemma}

\subsection{Proof of Claim~\ref{cla:tensor_trick}}\label{sec:notation_tools:tensor_tool}
In this section, we give the missing proof of Claim~\ref{cla:tensor_trick}.
\begin{proof}
 Let $H:=[h_1, \cdots, h_d]\in\R^{d\times d}$, then $h=\vect(H)=[h_1^\top, \cdots, h_d^\top]^\top$ and thus 
 \begin{align*}
     (V^\top H V )_{i,i} = & ~ v_i^\top H v_i \\
     = & ~ v_i^\top[h_1, \cdots, h_d]v_i \\
     = & ~ \sum_{j=1}^{d}v_{i,j}v_i^\top h_j \\
     = & ~ [v_{i,1}v_i^\top, \cdots, v_{i,d}v_i^\top ] [h_1^\top, \cdots, h_d^\top ]^\top\\
     = & ~ (\vect(v_i v_i^\top) )^\top\cdot h \\
     = & ~ (Uh)_{i},
 \end{align*}
 \
where the first step follows from the property of matrix multiplication: $(AB)_{i,j}=A_{i,:}B_{:,j}$; the third step follows from $v_i=[v_{i,1}, \cdots, v_{i,d}]^\top$; the last step follows from $U_{i,:}= (\vect(v_i v_i^\top) )^\top$.
\end{proof}

\subsection{Generalized Tensor Case}\label{sec:notation_tools_general_tensor_tools}
Consider the case $x_i=b_i\otimes a_i$ for some $a_i\in\R^p$, $b_i\in\R^q$, and $p\cdot q=d$. 
Similar to Fact~\ref{cla:tensor_trick}, we have the following statement. 
\begin{claim}\label{cla:general_tensor}
Let $A:=[a_1, \cdots, a_n]\in\R^{p\times n}$, $B:=[b_1, \cdots, b_n]\in\R^{q\times n}$, and $X:=[x_1, \cdots, x_n]\in\R^{d\times n}$ such that $x_i=b_i\otimes a_i$ for all $i\in[n]$, then for any $i,j\in[n]$, it holds that $ (A^\top\cdot\vect^{-1}(x_i)\cdot B )_{j,j}= (X^\top x_i )_j$. 
\end{claim}
\begin{proof}
We can show
\begin{align*}
\text{RHS}
= & ~ x_j^\top x_i \\
= & ~ (b_j\otimes a_j)^\top(b_i\otimes a_i) \\
= & ~ (b_j^\top\otimes a_j^\top )(b_i\otimes a_i) \\
= & ~ (b_j^\top b_i)\otimes (a_j^\top a_i) \\
= & ~ a_j^\top a_i b_i^\top b_j \\
= & ~ \text{LHS},     
\end{align*}
% \begin{align*}
% \text{RHS}
% = & ~ x_j^\top x_i = (b_j\otimes a_j)^\top(b_i\otimes a_i) 
% = (b_j^\top\otimes a_j^\top )(b_i\otimes a_i) \\
% = & ~ (b_j^\top b_i)\otimes (a_j^\top a_i)
% = a_j^\top a_i b_i^\top b_j
% = \text{LHS},     
% \end{align*}
where the third step follows from the fact $(A\otimes B)^\top=A^\top \otimes B^\top$; the fourth step follows from the fact $(A_1\otimes B_1)\cdot(A_2\otimes B_2)=(A_1\cdot A_2)\otimes(B_1\cdot B_2)$; the last step follows from the fact $x_i=\vect(a_i b_i^\top)$.
\end{proof}

\iffalse
\begin{lemma}[Chernoff bound \cite{c52}]\label{lem:chernoff}
Let $X = \sum_{i=1}^n X_i$, where $X_i=1$ with probability $p_i$ and $X_i = 0$ with probability $1-p_i$, and all $X_i$ are independent. Let $\mu = \E[X] = \sum_{i=1}^n p_i$. Then \\
1. $ \Pr[ X \geq (1+\delta) \mu ] \leq \exp ( - \delta^2 \mu / 3 ) $, $\forall \delta > 0$ ; \\
2. $ \Pr[ X \leq (1-\delta) \mu ] \leq \exp ( - \delta^2 \mu / 2 ) $, $\forall 0 < \delta < 1$. 
\end{lemma}
\fi
%\subsection{Technical claims} %%% old name
\section{Technical Preparations %(Missing proofs from Section~\ref{sec:quartic_suffices})
}\label{sec:missing_proof}

%\Zhao{You need to say each subsection is doing what.} \Junwei{Done}
This section is organized as follows: In Section~\ref{sec:bound_G_tr}, we calculate the upper bound of $\|G_{t,r}\|_2$. Next, we finish the proof of Claim~\ref{cla:inductive_claim} in Section~\ref{sec:proof_of_inductive_claim}. Then, We derive bounds for $C_1$, $C_2$, $C_3$, $C_4$ in Section~\ref{sec:bound_C_1}, \ref{sec:bound_C_2}, \ref{sec:proof_c3}, \ref{sec:proof_c4}.
\subsection{Upper Bound of \texorpdfstring{$\|G_{t,r}\|_2$}{}}
\label{sec:bound_G_tr}
%\begin{lemma}[Formal version of Lemma~\ref{lem:G_tr_l2}]\label{lem:G_tr_l2_formal}
%For any $t\ge 0$ and $r\in[m]$, 
%\begin{align*}
%    \|G_{t,r}\|_2\le\frac{n}{\sqrt{mB}}\cdot\|u(t)-y\|_2. 
%\end{align*} 
%\end{lemma}

%Before proving Theorem~\ref{thm:quartic}, we first bound $\|G_{t,r}\|_2$, whose proof is deferred to Appendix~\ref{sec:bound_G_tr}. 
Firstly, we find the upper bound of the norm of $G_{t,r}$.
\begin{lemma}\label{lem:G_tr_l2}
For any $t\ge 0$ and $r\in[m]$, 
\begin{align*}
    \|G_{t,r}\|_2\le\frac{n}{\sqrt{m\cdot S_{\batch}}}\cdot\|u(t)-y\|_2. 
\end{align*} 
\end{lemma}

\begin{proof}%[Proof of Lemma~\ref{lem:G_tr_l2}]
Recall the definition of $G_{t,r}$, 
\begin{align*}
    G_{t,r}=\frac{n}{|S_t|}\cdot\frac{1}{\sqrt{m}}\sum_{i\in S_t}(u_i(t)-y_i)\cdot a_r\cdot \mathbbm{1}(w_r(t)^\top x_i>\tau)\cdot x_i,
\end{align*}
we have 
\begin{align*}
    \|G_{t,r}\|_2&=\frac{n}{S_{\batch}}\cdot\frac{1}{\sqrt{m}}\|\sum_{i\in S_t}(u_i(t)-y_i)\cdot a_r \cdot \mathbbm{1}(w_r(t)^\top x_i>\tau)\cdot x_i\|_2\\
    &\le\frac{n}{S_{\batch}}\cdot\frac{1}{\sqrt{m}}\sum_{i\in S_t}\|(u_i(t)-y_i)\cdot a_r \cdot \mathbbm{1}(w_r(t)^\top x_i>\tau)\cdot x_i\|_2\\
    &=\frac{n}{S_{\batch}}\cdot\frac{1}{\sqrt{m}}\sum_{i\in S_t}|u_i(t)-y_i|\cdot|a_r|\cdot|\mathbbm{1}(w_r(t)^\top x_i>\tau)|\cdot\|x_i\|_2\\
    &\le\frac{n}{S_{\batch}}\cdot\frac{1}{\sqrt{m}}\sum_{i\in S_t}|u_i(t)-y_i|\\
    &\le\frac{n}{\sqrt{m}}\cdot\frac{1}{\sqrt{S_{\batch}}}\sqrt{\sum_{i\in S_t}(u_i(t)-y_i)^2}\\
    &\le\frac{n}{\sqrt{m\cdot S_{\batch}}}\cdot\|u(t)-y\|_2, 
\end{align*}
where the second step follows from triangle inequality; the fourth step follows from the facts that $a_r\in\{-1, +1\}$ and $\|x_i\|_2=1$; the fifth step follows from Cauchy-Schwarz inequality.
\end{proof}

%\subsection{Bound the Movement of Weight Vectors}\label{sec:movement_weight_vectors}
%\begin{lemma}[Bounding the movement of the weight vector, formal version of Lemma~\ref{lem:movement_weight}]\label{lem:movement_weight_formal}
%If Eq. \eqref{eq:quartic_condition} holds for $0\le i\le t$, then with probability at least $1-\rho$, it holds that for all $r\in [m]$, 
%\begin{align*}
%& ~ \| w_r(t+1) - w_r(0) \|_2 \\
%\leq & ~ \frac{2n}{\lambda\sqrt{mB}}\cdot\|u(0)-y\|_2\\
%:= & ~ \gamma.
%\end{align*}
%\end{lemma}

\subsection{Proof of Claim \ref{cla:inductive_claim}}\label{sec:proof_of_inductive_claim}
Then we finish Claim~\ref{cla:inductive_claim} deferred to here.
\begin{proof}
By the formulations of $v_{1,i}$ and $v_{2,i}$ (see Eq.~\eqref{eq:v_1i} and Eq.~\eqref{eq:v_2i}), we have that for each $i\in[n]$, 
\begin{align*}
    u_i(t+1)-u_i(t)=v_{1,i}+v_{2,i}. 
\end{align*}
which has the following vector form 
\begin{align}\label{eq:u_tplus1_ut_v1_v2}
u(t+1) - u(t) = v_1 + v_2 .
\end{align}

For $v_{1,i}$, it holds that 
\begin{align}
    v_{1,i} &= \frac{1}{ \sqrt{m} } \sum_{r \in V_i} a_r ( \phi_\tau ( w_r(t+1)^\top x_i ) - \phi_\tau( w_r(t)^\top x_i ) )\notag\\
    &=\frac{1}{\sqrt{m}}\sum_{r\in V_i}a_r((w_r(t+1)^{\top}x_i-\tau)\cdot\mathbbm{1}(w_r(t+1)^\top x_i>\tau)\notag\\
    &-(w_r(t)^\top x_i-\tau)\cdot\mathbbm{1}(w_r(t)^\top x_i>\tau)).\label{eq:v_1i_expand}
\end{align}
By Lemma~\ref{lem:movement_weight}, we can bound the movement of weight vectors, i.e., 
\begin{align*}
    \|w_r(t)-w_r(0)\|_2\le \gamma. 
\end{align*}
By the definition of the set $V_i$, for each $r\in V_i$, we have that 
\begin{align}\label{eq:trans_wt_tplus1_0}
    \mathbbm{1}(w_r(t+1)^\top x_i>\tau)=\mathbbm{1}(w_r(t)^\top x_i>\tau)=\mathbbm{1}(w_r(0)^\top x_i>\tau). 
\end{align}
Combining Eq.~\eqref{eq:v_1i_expand} and Eq.~\eqref{eq:trans_wt_tplus1_0}, we have 
\begin{align*}
    v_{1,i}&=\frac{1}{\sqrt{m}}\sum_{r\in V_i}a_r((w_r(t+1)-w_r(t))^\top x_i\cdot\mathbbm{1}(w_r(t)^\top x_i>\tau))\\
    &=-\frac{1}{\sqrt{m}}\sum_{r\in V_i}a_r\cdot\eta\cdot G_{t,r}^\top x_i\cdot\mathbbm{1}(w_r(t)^\top x_i>\tau)\\
    &=-\frac{1}{\sqrt{m}}\sum_{r\in V_i}a_r\cdot \eta(\frac{n}{|S_t|}\cdot\frac{1}{\sqrt{m}}\sum_{j\in S_t}(u_j(t)-y_j)\cdot a_r x_j\cdot\mathbbm{1}(w_r(t)^\top x_j>\tau))^\top x_i\\
    &\cdot\mathbbm{1}(w_r(t)^\top x_i>\tau)\\
    &=\frac{n}{S_{\batch}}\cdot\frac{\eta}{m}\sum_{r\in V_i}\sum_{j\in S_t}(y_j-u_j(t))\cdot x_i^\top x_j\cdot\mathbbm{1}(w_r(t)^\top x_i>\tau, w_r(t)^\top x_j>\tau)\\
    &=\frac{n}{S_{\batch}}\cdot\eta\sum_{j\in S_t}(y_j-u_j(t))\cdot\frac{1}{m}\sum_{r\in V_i}x_i^\top x_j\cdot\mathbbm{1}(w_r(t)^\top x_i>\tau, w_r(t)^\top x_j>\tau)\\
    &=\frac{n}{S_{\batch}}\cdot\eta\sum_{j\in S_t}(y_j-u_j(t))\cdot(H(t)_{i,j}-H(t)^{\perp}_{i,j}),
\end{align*}
where the second step follows from the formulation of $w_r(t+1)$ (see Eq.~\eqref{eq:SGD_iter}); the third step follows from the definition of $G_{t,r}$ (see Eq.~\eqref{eq:def_noisy_G}); the fourth step follows from the fact that $a_r$ is sampled from $\{-1, +1\}$; the last step follows from the definitions of $H(t)_{i,j}$ and $H(t)^{\perp}_{i,j}$ (see Eq.~\eqref{eq:H_t} and Eq.~\eqref{eq:H_perp_t}). Then the vector $v_1 \in \R^n$ can be formulated as 
\begin{align}\label{eq:rewrite_v1}
v_1 = & ~ \frac{n}{S_{\batch}} \cdot \eta  \cdot [ H(t) - H(t)^{\bot} ]_{:, S_t} \cdot [ y - u(t) ]_{S_t}\notag \\
= & ~ \eta  \cdot  ( H(t) - H(t)^{\bot}) \cdot D_{t}\cdot ( y - u(t) ),
\end{align}
where $D_{t} \in \R^{n \times n}$ is a diagonal sampling matrix such that the set of indices of nonzero entries is $S_t$ and each nonzero entry is equal to $\frac{n}{S_{\batch}}$.

Now we rewrite $\| y - u(t+1) \|_2^2$ as follows 
\begin{align}\label{eq:y_u_tplus1_expand}
& ~\| y - u(t+1) \|_2^2 \notag\\
= & ~ \| y - u(t) - ( u(t+1) - u(t) ) \|_2^2 \notag\\
= & ~ \| y - u(t) \|_2^2 - 2 ( y - u(t) )^\top  ( u(t+1) - u(t) ) + \| u (t+1) - u(t) \|_2^2 .
\end{align}
For the second term in the above equation, it holds that 
\begin{align}\label{eq:y_u_tplus1_second}
 & ~ ( y - u(t) )^\top ( u(t+1) - u(t) ) \notag\\
= & ~ ( y - u(t) )^\top ( v_1 + v_2 )  \notag\\
= & ~ ( y - u(t) )^\top v_1 + ( y - u(t) )^\top v_2 \notag\\
= & ~  \eta ( y - u(t) )^\top \cdot  H(t)  \cdot D_{t} \cdot  ( y - u (t) )\notag\\
-& ~\eta ( y - u(t) )^\top \cdot  H(t)^{\bot}   \cdot D_{t} \cdot  ( y - u(t) ) + ( y - u(t) )^\top v_2,
\end{align}
where the first step follows from Eq.~\eqref{eq:u_tplus1_ut_v1_v2}; the third step follows from Eq.~\eqref{eq:rewrite_v1}. Combining Eq.~\eqref{eq:y_u_tplus1_expand} and Eq.~\eqref{eq:y_u_tplus1_second}, and the definitions of quantities $C_1$, $C_2$, $C_3$, and $C_4$ (see Eq.~\eqref{def:C1}, \eqref{def:C2}, \eqref{def:C3}, and \eqref{def:C4}), we have
\begin{align*}
  \| y - u(t+1) \|_2^2  
=   \| y - u(t) \|_2^2 + C_1 + C_2 + C_3 + C_4 . 
\end{align*}
\end{proof}

%\Zhao{Once you change $0$ to $\tau$, the following lemma is NOT lemma 3.1 in \cite{dzps19}. You might need to cite some lemma in \cite{syz21}. If their lemma can be cited directly, you don't need to provide any proof, I think.}\Mingquan{Done.} \Zhao{I don't think it's done. This section is still confusing.}

%\Zhao{I don't think \cite{dzps19} is considering shift NTK. Their paper only consider $\tau=0$. Look at the following statement. You wrote general $\tau$. This cannot be true.}
%\begin{lemma}[Lemma 3.1 in \cite{dzps19}]\label{lem:3.1}
%We define $H^{\cts}, H^{\dis} \in \R^{n \times n}$ as follows
%\begin{align*}
%H^{\cts}_{i,j} = & ~ \E_{w \sim \N(0,I)} [ x_i^\top x_j {\bf 1}_{ w^\top x_i >\tau, w^\top x_j >\tau } ] , \\ 
%H^{\dis}_{i,j} = & ~ \frac{1}{m} \sum_{r=1}^m [ x_i^\top x_j {\bf 1}_{ w_r^\top x_i >\tau, w_r^\top x_j >\tau } ].
%\end{align*}
%Let $\lambda = \lambda_{\min} (H^{\cts}) $. If $m = \Omega( \lambda^{-2} n^2\log (n/\rho) )$, we have 
%\begin{align*}
%\| H^{\dis} - H^{\cts} \|_F \leq \frac{ \lambda }{4}, \mathrm{~and~} \lambda_{\min} ( H^{\dis} ) \geq \frac{3}{4} \lambda.
%\end{align*}
%hold with probability at least $1-\rho$.
%\end{lemma}
%The proof can be found in Appendix \ref{sec:lem_3.1}.

\subsection{Bound for \texorpdfstring{$C_1$}{}}\label{sec:bound_C_1}
We first introduce the following lemma which is necessary for bounding $C_1$. 

\begin{lemma}[Lemma C.2 in~\cite{syz21}]\label{lem:H0_Ht_diff}
Let $\tau>0$ and $\gamma\le 1/\tau$. Let $c, c'>0$ be two fixed constants. If $\|w_r(t)-w_r(0)\|_2\le\gamma$ holds for each $r\in[m]$, then 
\begin{align*}
    \|H(t)-H(0)\|_F\le n\cdot\min\{c\cdot\exp(-\tau^2/2), 3\gamma\}
\end{align*}
holds with probability at least $1-n^2\cdot\exp(-m\cdot\min\{c'\cdot\exp(-\tau^2/2), \gamma/10\})$. 
\end{lemma}
In addition, the random sampling set $S_t$ has the following fact. 
\begin{fact}\label{fact:E_Dt_I}
$\E_{S_t}[D_t]=I$.
\end{fact}
\begin{proof}
For each $i\in[n]$, 
\begin{align*}
    \E[(D_t)_{i,i}]=\frac{n}{S_{\batch}} \cdot \frac{{n-1 \choose S_{\batch}-1 }}{{n \choose S_{\batch}}} = 1.
\end{align*}
This completes the proof. 
\end{proof}
Now we give the bound for $C_1$. 
\begin{claim}\label{cla:C1}
Let $C_1 = -2 \eta (y - u(t))^\top \cdot H(t) \cdot  D_{t} \cdot ( y - u(t) )$, then 
\begin{align*}
\E_{S_t}[ C_1] \leq -2\eta\cdot(3\lambda/4-3n\gamma)\cdot\|y-u(t)\|_2^2
\end{align*}
holds with probability at least $1-(1/c+\rho+n^2\cdot\exp(-m\cdot\min\{c'\cdot\exp(-\tau^2/2), \gamma/10\})+\alpha)$. 
\end{claim}
\begin{proof}
\iffalse
By the definition of $C_1$, we have 
\begin{align*}
    C_1&=-2\eta(y-u(t))^\top\cdot H(t)\cdot D_t\cdot (y-u(t))\\
    &\le -2\eta\cdot \lambda_{\min}(H(t)\cdot D_t)(y-u(t))^\top (y-u(t))\\
    &=-\frac{2\eta n}{S_{\batch}}\cdot\lambda_{\min}(I_t\cdot H(t)\cdot I_t)(y-u(t))^\top(y-u(t))
\end{align*}
\Mingquan{I want to get rid of expectation $\E$ for $C_1$, $C_2$, $C_3$, $C_4$, and then Eq.~\eqref{eq:quartic_condition}, but it does not work.}
\fi

By the definition of $C_1$, we have 
\begin{align*}
     \E[C_1] = & ~ -2 \eta (y-u(t))^\top \cdot H(t) \cdot \E_{S_t}[D_{t}] \cdot ( y - u(t) ) \\
    = & ~ - 2 \eta (y-u(t))^\top \cdot H(t) \cdot \E_{S_t}[ D_{t}] \cdot ( y - u(t) ) \\
    = & ~ - 2 \eta (y-u(t))^\top \cdot H(t) \cdot I \cdot ( y - u(t) ),
\end{align*}
where the third step follows from Fact~\ref{fact:E_Dt_I}. 

Lemma~\ref{lem:movement_weight} gives us that with probability at least $(1-1/c)\cdot(1-\rho)$ 
for all $r\in[m]$, 
\begin{align}\label{eq:w_rt_w_r0_diff}
    \|w_r(t)-w_r(0)\|_2\le\gamma. 
\end{align}
Combining Eq.~\eqref{eq:w_rt_w_r0_diff} and Lemma \ref{lem:H0_Ht_diff}, we have with probability at least $1-n^2\cdot\exp(-m\cdot\min\{c'\cdot\exp(-\tau^2/2), \gamma/10\})$,
\begin{align}\label{eq:H0_Ht_F}
\|H(0)-H(t)\|_F\leq 3n\gamma.    
\end{align}
Moreover, it holds that 
\begin{align}\label{eq:H0_Ht_2}
    \|H(0)-H(t)\|_2\ge \lambda_{\max}(H(0)-H(t))\ge\lambda_{\min}(H(0))-\lambda_{\min}(H(t)).
\end{align}
Note that $H(0)=H^{\dis}$, by Lemma~\ref{lem:H_dis_lambda_min}, with probability at least $1-\alpha$,
\begin{align}\label{eq:H0_lambda_min}
\lambda_{\min}(H(0))\ge\frac{3}{4}\lambda.     
\end{align}
Then Eq.~\eqref{eq:H0_Ht_2} becomes 
\begin{align*}
\lambda_{\min}(H(t)) &\geq \lambda_{\min}(H(0))- \|H(0)-H(t)\|_2\\
&\ge \lambda_{\min}(H(0))-\|H(0)-H(t)\|_F\\
&\geq 3\lambda/4-3n\gamma,
\end{align*}
where the second step follows from $\|H(0)-H(t)\|_2\le\|H(0)-H(t)\|_F$; the third step follows from Eq.~\eqref{eq:H0_lambda_min} and Eq.~\eqref{eq:H0_Ht_F}. Therefore, we have 
\begin{align*}
  (y - u(t))^\top H(t) ( y - u(t) ) \geq(3\lambda/4-3n\gamma)\cdot\| y - u(t) \|_2^2
\end{align*}
with probability at least $1-(1/c+\rho+n^2\cdot\exp(-m\cdot\min\{c'\cdot\exp(-\tau^2/2), \gamma/10\})+\alpha)$ by union bound. 
\end{proof}

\subsection{Bound for \texorpdfstring{$C_2$}{}}\label{sec:bound_C_2}
Before bounding $C_2$, we present the following claim. 
\begin{claim}[Claim C.11 in \cite{syz21}]\label{cla:r_bar_Vi_Pr}
Let $\gamma\le\frac{1}{\tau}$, then for each $r\in[m]$, 
\begin{align*}
    \Pr[r\in\ov{V}_i]\le\min\{\gamma, O(\exp(-\tau^2/2))\}.
\end{align*}
\end{claim}
The following fact gives the upper bound of $\|H(t)^\top\|_F^2$, which is used in the following proof. 
\begin{fact}\label{fact:bound_H_k_bot}
\begin{align*}
\| H(t)^{\bot} \|_F^2 \leq \frac{n}{m^2} \sum_{i=1}^n ( \sum_{r=1}^m \mathbbm{1}(r\in \ov{V}_i))^2.
\end{align*}
\end{fact}

\begin{proof}
We have
\begin{align*}
\| H(t)^{\bot} \|_F^2
= & ~ \sum_{i=1}^n\sum_{j=1}^n (H(t)^{\bot}_{i,j})^2\\
= & ~ \sum_{i=1}^n\sum_{j=1}^n ( \frac {1} {m}\sum_{r\in \ov{V}_i} x_i^\top x_j\cdot\mathbbm{1}(w_r(t)^\top x_i>\tau, w_r(t)^\top x_j>\tau))^2\\
= & ~ \sum_{i=1}^n\sum_{j=1}^n ( \frac {1} {m}\sum_{r=1}^m x_i^\top x_j\cdot\mathbbm{1}(w_r(t)^\top x_i>\tau, w_r(t)^\top x_j>\tau)\cdot \mathbbm{1}(r\in \ov{V}_i))^2\\
= & ~ \sum_{i=1}^n\sum_{j=1}^n ( \frac {x_i^\top x_j} {m} )^2 ( \sum_{r=1}^m \mathbbm{1}(w_r(t)^\top x_i>\tau, w_r(t)^\top x_j>\tau) \cdot \mathbbm{1}(r\in \ov{V}_i))^2 \\
\leq & ~ \frac{1}{m^2} \sum_{i=1}^n\sum_{j=1}^n ( \sum_{r=1}^m \mathbbm{1}(w_r(t)^\top x_i>\tau, w_r(t)^\top x_j>\tau)\cdot \mathbbm{1}(r\in \ov{V}_i) )^2 \\
= & ~ \frac{n}{m^2} \sum_{i=1}^n ( \sum_{r=1}^m \mathbbm{1}(r\in \ov{V}_i) )^2.
\end{align*}
\end{proof}

\iffalse
\subsection{Proof of Fact \ref{fact:dudt} \Zhao{This label is missing.} }\label{sec:dudt_proof}
\begin{proof}
For each $i\in [n]$,
we have

\begin{align*}
&~\frac{\d}{\d t} u_i(t)\\
=&~\sum_{r=1}^m \langle \frac{\partial f(W(t),a,x_i)}{\partial w_r(t)},\frac{\d w_r(t)}{\d t} \rangle\\
=&~\sum_{r=1}^m \langle \frac{\partial f(W(t),a,x_i)}{\partial w_r(t)},-\frac{\partial L(W(t),a)}{\partial w_r(t)} \rangle\\
 =&~\sum_{r=1}^m \langle \frac{\partial f(W(t),a,x_i)}{\partial w_r(t)},
 -\frac{1}{ \sqrt{m} } \sum_{i=1}^n ( f(W(t),x_i,a_r) - y_i ) a_r x_i {\bf 1}_{ w_r(t)^\top x_i >\tau } \rangle\\
=&~\sum_{j=1}^n (y_j-u_j(t)) \langle \frac{\partial f(W(t),a,x_i)}{\partial w_r(t)},\frac{\partial f(W(t),a,x_j)}{\partial w_r(t)} \rangle\\
=&~\sum_{j=1}^n (y_j-u_j(t))H(t)_{i,j}
\end{align*}
where the first step follows from Eq. \eqref{eq:ut_def} and the chain rule of derivatives,
the second step uses Eq. \eqref{eq:gradient},
the third step uses Eq. \eqref{eq:wr_derivative},
the fourth step uses Eq. \eqref{eq:relu_derivative} and Eq. \eqref{eq:ut_def},
and the last step uses the definition of the matrix $H$.
\end{proof}
\fi
Now we give the bound for $C_2$. 
\begin{claim}\label{cla:C2}
Let $C_2 = 2 \eta ( y - u(t) )^\top \cdot H(t)^{\bot} \cdot  D_{t}  \cdot ( y - u(t) )$, then
\begin{align*}
\E_{S_t}[ C_2 ] \leq 6n\eta\gamma\cdot \| y - u(t) \|_2^2
\end{align*}
holds with probability $1-n\cdot \exp(-9m\gamma/4)$.
\end{claim}
\begin{proof}
Similarly as before, we have
\begin{align*}
    \E[C_2] = & ~ \E_{S_t} [ 2 \eta \cdot (y-u(t))^\top \cdot H(t)^{\bot} \cdot D_{t} \cdot (y-u(t)) ] \\
    = & ~  2 \eta \cdot (y-u(t))^\top \cdot H(t)^{\bot} \cdot \E_{S_t} [ D_{t} ] \cdot (y-u(t))  \\
    = & ~  2 \eta \cdot (y-u(t))^\top \cdot H(t)^{\bot} \cdot I \cdot (y-u(t)),
\end{align*}
where the last step follows from Fact~\ref{fact:E_Dt_I}. Furthermore,
\begin{align*}
\E[ C_2 ] \leq 2 \eta\cdot \| H(t)^{\bot} \|_2\cdot \| y - u(t)\|_2^2 .
\end{align*}
Therefore, it suffices to upper bound $\| H(t)^{\bot} \|_2$. 

For each $i\in[n]$, we define $Z_i:=\sum_{r=1}^{m}\mathbbm{1}(r\in\ov{V}_i)$. Note that the $m$ random variables $\{\mathbbm{1}(r\in\ov{V}_i)\}_{r=1}^{m}$ are mutually independent since $\mathbbm{1}(r\in\ov{V}_i)$ only depends on $w_r(0)$. In addition, for each $r\in[m]$, it trivially holds that $|\mathbbm{1}(r\in\ov{V}_i)|\le 1$. By Claim~\ref{cla:r_bar_Vi_Pr}, we have $\E[\mathbbm{1}(r\in\ov{V}_i)]\le\gamma$. In particular, 
\begin{align}\label{eq:E_square_am_R}
    \E[\mathbbm{1}(r\in\ov{V}_i)^2]=\E[\mathbbm{1}(r\in\ov{V}_i)]\le\gamma. 
\end{align}
Applying Bernstein inequality (see Lemma~\ref{lem:bernstein}) gives us 
\begin{align*}
    \Pr[Z_i\ge a]&\le\exp(-\frac{a^2/2}{\sum_{r=1}^{m}\E[\mathbbm{1}(r\in\ov{V}_i)^2]+a/3})\\
    &\le\exp(-\frac{a^2/2}{mR+a/3}),
\end{align*}
where the last step follows from Eq.~\eqref{eq:E_square_am_R}. Setting $a=3m\gamma$, we have 
\begin{align*}
    \Pr[Z_i\ge 3m\gamma]\le\exp(-9m\gamma/4). 
\end{align*}
Moreover, by union bound, we have that
\begin{align*}
    \forall i\in[n],\ Z_i\le 3m\gamma, 
\end{align*}
with probability at least $1-n\cdot\exp(-9m\gamma/4)$. 

By Fact~\ref{fact:bound_H_k_bot}, we know that 
\begin{align*}
    \|H(t)^{\perp}\|_F^2&\le\frac{n}{m^2}\sum_{i=1}^{n}(\sum_{r=1}^{m}\mathbbm{1}(r\in\ov{V}_i))^2\\
    &=\frac{n}{m^2}\sum_{i=1}^{n}Z_i^2\\
    &\le 9n^2 \gamma^2,
\end{align*}
where the second step follows from the definition of $Z_i$; the third step follows with probability at least $1-n\cdot\exp(-9m\gamma/4)$. Furthermore, we have 
\begin{align*}
    \|H(t)^{\perp}\|_2\le\|H(t)^{\perp}\|_F\le 3n\gamma
\end{align*}
with probability at least $1-n\cdot\exp(-9m\gamma/4)$. Then we have
\begin{align*}
    \E[C_2]\le 6n\eta\gamma\cdot \|y-u(t)\|_2^2. 
\end{align*}

This completes the proof. 

\end{proof}
\subsection{Bound for \texorpdfstring{$C_3$}{}}\label{sec:proof_c3}
In this subsection, we derive the upper bound for $C_3$ by bounding $\| \E[ v_2 ] \|_2$. 
\begin{claim}\label{cla:C3}
Let $C_3 = - 2 (y - u(t))^\top v_2$, then 
\begin{align*}
\E[ C_3 ] \leq \frac{6n^{3/2}\eta\gamma}{\sqrt{S_{\batch}}}\cdot\|y-u(t)\|_2^2
\end{align*}
with probability at least $1-n\cdot\exp(-9m\gamma/4)$.
\end{claim}
\begin{proof}
Using Cauchy-Schwarz inequality, we have
\begin{align*}
\E[ C_3]
= & ~- \E[ 2(y-u(t))^\top v_2 ] \\
= & ~ -2(y-u(t))^\top \E[v_2] \\
\leq & ~ 2 \| y - u(t) \|_2 \cdot \| \E[ v_2 ] \|_2. 
\end{align*}

We can upper bound $\| \E[ v_2 ] \|_2$ in the following sense
\begin{align*}
\| \E[ v_2 ] \|_2^2
\leq &~ \sum_{i=1}^n (\frac{\eta}{ \sqrt{m} } \sum_{ r \in \ov{V}_i } | {G}_{t,r}^\top x_i |)^2\\
= &~ \frac{\eta^2}{ m }\sum_{i=1}^n (\sum_{r=1}^m \mathbbm{1}(r\in \ov{V}_i)\cdot| {G}_{t,r}^\top x_i |)^2\\
\leq &~ \frac{\eta^2}{ m }\cdot \max_{r \in [m]} \| {G}_{t,r} \|_2^2\cdot\sum_{i=1}^n (\sum_{r=1}^m \mathbbm{1}(r\in \ov{V}_i))^2\\
 \leq & ~ \frac{\eta^2}{ m }\cdot (\frac{n}{ \sqrt{m\cdot S_{\batch}} }\cdot\| u(t) - y\|_2 )^2 \cdot \sum_{i=1}^n (\sum_{r=1}^m \mathbbm{1}(r\in \ov{V}_i))^2\\
  \leq & ~ \frac{\eta^2}{ m }\cdot (\frac{n}{ \sqrt{m\cdot S_{\batch}} }\cdot\| u(t) - y\|_2 )^2 \cdot \sum_{i=1}^n (3m\gamma)^2\\
  = & ~ \frac{9\eta^2 n^3 \gamma^2}{S_{\batch}}\cdot \| u(t) - y\|_2^2,
\end{align*}
where the first step follows from the definition of $v_2$ (see Eq.~\eqref{eq:v_2i}) and the property of function $\phi_{\tau}$; the third step follows from Cauchy-Schwarz inequality and the fact that $\|x_i\|_2=1$; the fourth step follows from Lemma~\ref{lem:G_tr_l2}; the fifth step follows from the fact that $\sum_{r=1}^m \mathbbm{1}(r \in \ov{V}_i) \leq 3 m \gamma$ holds with probability at least $1-n\cdot\exp(-9m\gamma/4)$ which is proven in Claim~\ref{cla:C2}.
\end{proof}

\subsection{Bound for \texorpdfstring{$C_4$}{}}\label{sec:proof_c4}
Now, we calculate the upper bound of $C_4$ to prepare for the proof of convergence in Section~\ref{app:proof_convergence}.
\begin{claim}\label{cla:C4}
Let $C_4 = \| u(t+1) - u(t) \|_2^2$, then 
\begin{align*}
C_4 \leq \frac{n^3\eta^2}{S_{\batch}} \cdot \| y - u(t) \|_2^2.
\end{align*}
\end{claim}
\begin{proof}
We need to upper bound
\begin{align*}
    & ~  \| u(t+1) - u(t) \|_2^2    \\
    = & ~   \sum_{i=1}^n ( \frac{1}{\sqrt{m}} \sum_{r=1}^m a_r \cdot ( \phi_\tau ( ( w_r(t) - \eta \cdot {G}_{t,r} )^\top x_i ) - \phi_\tau ( w_r(t)^\top x_i ) ) )^2  \\
    \leq & ~   \sum_{i=1}^n ( \frac{1}{\sqrt{m}} \sum_{r=1}^m | \eta \cdot {G}_{t,r}^\top  \cdot x_i | )^2  \\
    \leq & ~ \eta^2 n \cdot\frac{1}{m} \cdot ( \sum_{r=1}^m \| {G}_{t,r} \|_2 )^2   \\
    \leq & ~ \frac{n^3\eta^2}{S_{\batch}} \cdot \| y - u(t) \|_2^2,
\end{align*}
where the first step follows from the definition of $w_r(t+1)$; the second step follows from the property of shifted ReLU and $a_r\in\{-1, +1\}$; the fourth step follows from triangle inequality; the last step follows from Lemma~\ref{lem:G_tr_l2}.
\end{proof}
\section{Proof of Convergence}\label{app:proof_convergence}
In this section, we give the proof of Theorem~\ref{thm:quartic}. The proof mainly consists of two parts: (1)\ showing that the weight vector $w_r$ with $r\in[m]$ does not move too far from initialization; (2)\ showing that as long as the weight vector does not change too much, then the error $\|u(t)-y\|_2$ decays linearly with extra error term. We proceed the proof via a double induction argument, in which we assume these two conditions hold up to iteration $t$ and prove that they also hold simultaneously for iteration $t+1$.

We prove Theorem~\ref{thm:quartic} by induction.
The base case is $i=0$ and it is trivially true.
Assume that Eq.~\eqref{eq:quartic_condition} is true
for $0\le i\le t$, then our goal is to prove that Eq. \eqref{eq:quartic_condition} also holds for $i=t+1$.

From the induction hypothesis, we have the following lemma which states that the weight vectors should not change too much. 
\begin{lemma}\label{lem:movement_weight}
If Eq. \eqref{eq:quartic_condition} holds for $0\le i\le t$, then with probability at least $(1-1/c)\cdot(1-\rho)$ where $c>1$, it holds that for all $r\in [m]$, 
\begin{align*}
\| w_r(t+1) - w_r(0) \|_2\leq \frac{2\sqrt{c}n}{\lambda\sqrt{m\cdot S_{\batch}}}\cdot\|u(0)-y\|_2\le\gamma,
\end{align*}
where the parameter $\gamma$ is determined later.
\end{lemma}

\begin{proof}%[Proof of Lemma~\ref{lem:movement_weight}]
We first define the two events $\mathcal{E}_1$ and $\mathcal{E}_2$ to be 
\begin{align*}
    \mathcal{E}_1: &~\text{Eq.~\eqref{eq:quartic_condition} holds for $0\le i\le t$},\\
    \mathcal{E}_2: &~\text{$\|u(t)-y\|_2^2\le c\cdot(1-\eta\lambda/2)^t\cdot\|u(0)-y\|_2^2$ holds for $0\le i\le t$},
\end{align*}
where $c>1$ is a constant. By Markov's inequality, we have $\Pr[\mathcal{E}_2\mid\mathcal{E}_1]\ge 1-1/c$. Furthermore, it holds that 
\begin{align}\label{eq:Pr_E2}
    \Pr[\mathcal{E}_2]\ge\Pr[\mathcal{E}_2\mid\mathcal{E}_1]\cdot\Pr[\mathcal{E}_1]\ge (1-1/c)\cdot(1-\rho). 
\end{align}

For $t+1$, we have 
\begin{align*}
    \|w_r(t+1)-w_r(0)\|_2&=\|\eta\sum_{i=0}^{t}G_{i,r}\|_2\\
    &\le\eta\sum_{i=0}^{t}\|G_{i,r}\|_2\\
    &\le\eta\sum_{i=0}^{t}\frac{n}{\sqrt{m\cdot S_{\batch}}}\cdot\|u(i)-y\|_2\\
    &\le\frac{\sqrt{c}\eta n}{\sqrt{m\cdot S_{\batch}}}\sum_{i=0}^{t}(1-\eta\lambda/2)^{i/2}\cdot\|u(0)-y\|_2\\
    &\le\frac{2\sqrt{c}n}{\lambda\sqrt{m\cdot S_{\batch}}}\cdot\|u(0)-y\|_2, 
\end{align*}
where the first step follows from Eq.~\eqref{eq:SGD_iter} and  Eq.~\eqref{eq:def_noisy_G}; the second step follows from triangle inequality; the third step follows from Lemma~\ref{lem:G_tr_l2}; the fourth step follows from Eq.~\eqref{eq:Pr_E2}; the last step follows from the truncated geometric series.
%\Mingquan{In the fourth step, the induction hypothesis is $\E[\|u(t)-y\|_2^2]\le$, but we need $\|u(t)-y\|_2\le$.} \Zhao{I think we do conditional expectation}\Mingquan{How to do that?}
\end{proof}

For the initial error $\|u(0)-y\|_2$, we have the following claim. 
 \begin{claim}[Claim D.1 in~\cite{syz21}]
 \label{cla:yu0}
For $\beta\in(0, 1)$,
with probability at least $1-\beta$,
\begin{align*}
\|y-u(0)\|_2^2=O(n(1+\tau^2)\log^2(n/\beta)).
\end{align*}
\end{claim}

Next, we calculate the difference of predictions between two consecutive iterations. For each $i \in [n]$, we have
\begin{align*}
& ~ u_i(t+1) - u_i(t) \\
= & ~ \frac{1}{ \sqrt{m} } \sum_{r=1}^m a_r \cdot ( \phi_\tau( w_r(t+1)^\top x_i ) - \phi_\tau(w_r(t)^\top x_i ) ) \\
= & ~ \frac{1}{\sqrt{m}} \sum_{r=1}^m a_r \cdot ( \phi_\tau ( \big( w_r(t) - \eta \cdot G_{t,r}  \big)^\top x_i ) -  \phi_\tau ( w_r(t)^\top x_i ) ).
\end{align*}

The right hand side can be divided into two parts: $v_{1,i}$ represents one term that does not change and $v_{2,i}$ represents one term that may change. For each $i \in [n]$,
we define the set $V_i\subset [m]$ by 
\begin{align*}
    V_i:=\{r\in [m]:\forall 
    w\in \mathbb{R}^d \text{ such that } \|w-w_r(0)\|_2\leq \gamma,\ \mathbbm{1}(w_r(0)^\top x_i>\tau)=\mathbbm{1}(w^\top x_i>\tau)\},
\end{align*}
and $\overline{V}_i:=[m]\setminus V_i$. Then the quantities $v_{1,i}$ and $v_{2,i}$ are defined as follows
\begin{align}
v_{1,i} : = & ~ \frac{1}{ \sqrt{m} } \sum_{r \in V_i} a_r ( \phi_\tau ( \big( w_r(t) - \eta \cdot  {G}_{t,r}  \big)^\top x_i ) - \phi_\tau( w_r(t)^\top x_i ) ), \label{eq:v_1i} \\
v_{2,i} : = & ~ \frac{1}{ \sqrt{m} } \sum_{r \in \overline{V}_i } a_r ( \phi_\tau ( \big( w_r(t) - \eta \cdot  {G}_{t,r} \big)^\top x_i ) - \phi_\tau( w_r(t)^\top x_i ) ).\label{eq:v_2i}
\end{align} 

Given the definition of matrix $H(t)\in\R^{n\times n}$ such that \begin{align}
H(t)_{i,j} :=  ~ \frac{1}{m} \sum_{r=1}^m x_i^\top x_j\cdot\mathbbm{1}( w_r(t)^\top x_i >\tau, w_r(t)^\top x_j>\tau), \label{eq:H_t}
\end{align}
we define the matrix $H(t)^{\perp}\in\R^{n\times n}$ such that 
\begin{align}
H(t)^{\bot}_{i,j} :=  ~ \frac{1}{m} \sum_{r\in \ov{V}_i} x_i^\top x_j\cdot\mathbbm{1}( w_r(t)^\top x_i>\tau, w_r(t)^\top x_j>\tau ).\label{eq:H_perp_t}
\end{align}

Given $H(t), H(t)^{\perp}\in\R^{n\times n}$, we need the following four quantities which are components of $\|y-u(t+1)\|_2^2$. 
\begin{definition}
Define the quantities $C_1$, $C_2$, $C_3$, and $C_4$ by  
\begin{align}
C_1 := & ~ -2 \eta (y - u(t))^\top H(t) \cdot D_{t} \cdot ( y - u(t) ) , \label{def:C1}\\
C_2 := & ~ + 2 \eta ( y - u(t) )^\top H(t)^{\bot} \cdot D_{t} \cdot ( y - u(t) ) ,\label{def:C2} \\
C_3 := & ~ - 2 ( y - u(t) )^\top v_2 ,\label{def:C3} \\
C_4 := & ~ \| u (t+1) - u(t) \|_2^2,  \label{def:C4}
\end{align}
where $D_{t} \in \R^{n \times n}$ is a diagonal sampling matrix such that the set of indices of the nonzero entries is $S_t$ and each nonzero entry is equal to $\frac{n}{B}$. \iffalse
Moreover, we can rewrite $D_t$ by
\begin{align}
    D_t=\frac{n}{B}\cdot I_t,
\end{align}
where $I_t^2=I_t$. 
\fi %%%Zhao: I moved it to here.
\end{definition}
%%%\fi %%%Zhao: originally you put fi here
Now we can decompose the error term $\|y-u(t+1)\|_2^2$ into the following components and bound them later. 
\begin{claim}\label{cla:inductive_claim}
The difference between $u(t+1)$ and $y$ can be formulated as 
\begin{align*}
 \| y - u(t+1) \|_2^2  = \| y - u(t) \|_2^2 + C_1 + C_2 + C_3 + C_4.
\end{align*}
\end{claim}
The proof for Claim~\ref{cla:inductive_claim} is deferred to Appendix~\ref{sec:proof_of_inductive_claim}.

Armed with the above statements, now we prove the convergence theorem. For the sake of completeness, we include Theorem~\ref{thm:quartic} below. 
\begin{theorem}[Restatement of Theorem~\ref{thm:quartic}]
Given $n$ training samples $\{(x_i, y_i)\}_{i=1}^{n}$ and a parameter $\rho\in(0,1)$. Initialize $w_r\sim\N(0, I_d)$ and sample $a_r$ from $\{-1,+1\}$ uniformly at random for each $r\in[m]$. Set the width of neural network to be 
\begin{align*}
    m=\poly(\lambda^{-1}, S_{\batch}^{-1}, n, \log(n/\rho)),
\end{align*}
and the step size $\eta=\poly(\lambda, S_{\batch}, n^{-1})$, where $\lambda=\lambda_{\min}(H^{\cts})$ and $S_{\batch}$ is the batch size, then with probability at least $1-O(\rho)$, the vector $u(t)$ for $t\ge 0$ in Algorithm~\ref{alg:sgd_at} satisfies that 
\begin{align}\label{eq:quartic_bound}
  \E[\| u (t) - y \|_2^2]  \leq ( 1 - \eta \lambda / 2 )^t \cdot \| u (0) - y \|_2^2 .
\end{align}
\end{theorem}
\begin{proof}
By the linearity of expectation, applying Claim~\ref{cla:C1}, \ref{cla:C2}, \ref{cla:C3}, and \ref{cla:C4} gives us 
\begin{align*}
     &\E[\|y-u(t+1)\|_2^2]\\
     =&\|y-u(t)\|_2^2+\E[C_1]+\E[C_2]+\E[C_3]+\E[C_4]\\
     \le &(1-2\eta(3\lambda/4-3n\gamma)+6n\eta\gamma+6n^{3/2}\eta\gamma/\sqrt{S_{\batch}}+n^3\eta^2/S_{\batch})\cdot\|y-u(t)\|_2^2.
\end{align*}
\paragraph{Parameter Settings.} In order to satisfy Eq.~\eqref{eq:quartic_bound} for iteration $t+1$, let 
\begin{align}\label{eq:iter_ratio}
    1-2\eta(3\lambda/4-3n\gamma)+6n\eta\gamma+6n^{3/2}\eta\gamma/\sqrt{S_{\batch}}+n^3\eta^2/S_{\batch}\le 1-\eta\lambda/2. 
\end{align}
For the probability, we have 
\begin{align}\label{eq:prob_bound}
1/c+n^2\cdot\exp(-m\cdot\min\{c'\cdot\exp(-\tau^2/2), \gamma/10\})+\alpha+2n\cdot\exp(-9m\gamma/4)=O(\rho).
\end{align}

Lemma~\ref{lem:H0_Ht_diff} and Claim~\ref{cla:r_bar_Vi_Pr} require that
\begin{align}\label{eq:gamma_1_tau}
    \frac{2\sqrt{c}n}{\lambda\sqrt{m\cdot S_{\batch}}}\cdot\|u(0)-y\|_2\le\gamma\le 1/\tau,
\end{align}
where Claim~\ref{cla:yu0} gives that with probability at least $1-\beta$, 
\begin{align}\label{eq:y_u0_sq}
    \|y-u(0)\|_2^2=O(n(1+\tau^2)\log^2(n/\beta)).
\end{align}

Eq.~\eqref{eq:iter_ratio} implies that the step size $\eta$ satisfies that 
\begin{align*}
    \eta=O(\frac{\lambda\cdot S_{\batch}}{n^3})
\end{align*}
and $\gamma$ satisfies that 
\begin{align}\label{eq:gamma_value}
    \gamma=O(\frac{\lambda}{n}). 
\end{align}

By setting $\tau=O(\sqrt{\log{m}})$ and combining Eq.~\eqref{eq:gamma_1_tau}, \eqref{eq:y_u0_sq} and \eqref{eq:gamma_value}, we have 
\begin{align*}
    m=\wt{\Omega}(\frac{n^5}{\lambda^4\cdot S_{\batch}}).
\end{align*}
Taking the probability parameter $\rho$ in Eq.~\eqref{eq:prob_bound} into consideration, we have that 
\begin{align}
    m=\wt{\Omega}(\frac{n^5\cdot\log^{C}(n/\rho)}{\lambda^4\cdot S_{\batch}}),
\end{align}
where $C>0$ is a constant and the notation $\wt{\Omega}(\cdot)$ hides the factors $\log{m}$ and $\log{n}$. 

Thus, we complete the proof of Theorem~\ref{thm:quartic}. 
\end{proof}
\section{Asynchronous Tree Data Structure}
\label{sec:asynchronize_tree}
In this section, we present the \textsc{Init} operation of asynchronous tree data structure in Algorithm~\ref{alg:at_init}.  \textsc{Update} and \textsc{Query} have been shown in  Algorithm~\ref{alg:at_update} and Algorithm~\ref{alg:at_query} in Section~\ref{sec:technical_overview} respectively. Then, we give a theorem discussing the time complexity of asynchronous tree data structure.
\begin{itemize}
    \item The \textsc{Init} procedure constructs $n$ trees $T_1, \cdots, T_n$ for the $n$ data points $x_1, \cdots, x_n$. The tree $T_i$ with $i\in[n]$ has $m$ leaf nodes and the $r$-th leaf node with $r\in[m]$ has value $w_r^\top x_i$. The value of each inner node of $T_i$ is the maximum of the values of its two children. 
    \item The \textsc{Update} procedure updates the $n$ trees since the weight vector $w_r$ changes by $\delta_{t,r}$. It starts with the $r$-th leaf node of each $T_i$ whose value is added by $\delta_{t,r}^\top x_i$, and backtracks until to the root of $T_i$.
    \item The \textsc{Query} procedure returns the set of activated neurons for data point $x_i$ by searching the values in tree $T_i$ recursively from the root of $T_i$. 
\end{itemize}
For the asynchronous tree data structure, we have the following theorem and its proof is omitted. Note that the time complexity given in Theorem~\ref{thm:correlation_tree_data_structure} is for the general case, i.e., the input data has no special structures. When the data points have Kronecker structure, the time complexity for the initialization step and each iteration can be significantly accelerated.  
%\Zhao{There is no english here.}\Mingquan{Done.}
\begin{theorem}[\textsc{AsynchronousTree} data structure]\label{thm:correlation_tree_data_structure}
There exists a data structure with the following procedures:
\begin{itemize}
    \item \textsc{Init}$(\{w_1, \cdots, w_m\} \subset \R^d, \{x_1, \cdots, x_n\} \subset \R^d,n\in\mathbb{N},m\in\mathbb{N},d\in\mathbb{N})$. Given a series of weight vectors $w_1, \cdots, w_m$ and data vectors $x_1, \cdots, x_n$ in $d$-dimensional space, the preprocessing step takes time $O(n\cdot m\cdot d)$. 
    \item \textsc{Update}$(z\in\R^d,r\in [m])$. Given a vector $z$ and index $r$, it updates weight vector $w_r$ with $z$ in time $O(n\cdot(d+\log m))$.  
    \item \textsc{Query}$(i \in [n],\tau \in \R)$. Given an index $i$ corresponding to point $x_i$ and a threshold $\tau$, it finds all index $r\in[m]$ such that $w_r^\top x_i>\tau$ in time $O(|\tilde{S}(\tau)|\cdot \log m)$, where $\tilde{S}(\tau):=\{r: w_r^\top x_i>\tau\}$.
\end{itemize}
\end{theorem} 

\begin{algorithm}[!ht]\caption{Asynchronous tree data structure} \label{alg:at_init} 
\begin{algorithmic}[1]
\State {\bf data structure} \textsc{AsynchronousTree} 
\State {\bf members}
\State \hspace{4mm} $w_1, \cdots, w_m\in\R^d$\Comment{$m$ weight vectors}
\State \hspace{4mm} $x_1, \cdots, x_n \in \R^d$\Comment{$n$ data points}
\State \hspace{4mm} Binary trees $T_1, \cdots, T_n$ \Comment{$n$ binary search trees}
\State {\bf end members}
\State
\State {\bf public:}
\Procedure{Init}{$w_1, \cdots, w_m \in \R^d, x_1, \cdots, x_n \in \R^d$, $n$, $m$,  $d$}  \Comment{\textsc{Init} in Theorem~\ref{thm:correlation_tree_data_structure}}
    \For{$i=1 \to n$} \label{lin:init_first_loop}
        \State $x_i \gets x_i$
    \EndFor
    \For{$j=1 \to m$} \label{lin:init_second_loop}
        \State $w_j \gets w_j$
    \EndFor
    \For{$i=1 \to n$} \Comment{Each data point $x_i$ has a tree $T_i$} \label{lin:init_outer_loop}
        \For{$j=1 \to m$} \label{lin:init_inner_loop}
            \State $u_j \gets w_j^\top x_i$ \label{lin:init_inner_product}
        \EndFor
        \State $T_i \gets \textsc{MakeTree}(u_1, \cdots, u_m)$\label{lin:init_make_binary_tree} \Comment{Each node stores the maximum value of its two children}
    \EndFor
\EndProcedure
\State {\bf end data structure}
\end{algorithmic}
\end{algorithm}

\begin{algorithm}[!ht]\caption{Asynchronous tree data structure, restatement of Algorithm~\ref{alg:at_update_informal}}\label{alg:at_update}
\begin{algorithmic}[1]
\State {\bf data structure} \textsc{AsynchronousTree} 
\State {\bf public:}
\Procedure{Update}{$z\in\R^d, r \in [m]$} \Comment{\textsc{Update} in  Theorem~\ref{thm:correlation_tree_data_structure}}
\For{$i=1$ to $n$} \label{lin:update_loop}
    \State $l \gets$ the $r$-th leaf of tree $T_i$ \label{lin:update_find_leaf}
    \State $l.\text{value} \gets l.\text{value}+z^\top x_i$ \label{lin:update_inner_product}
    \While{$l$ is not root}
        \State $p$ $\gets$ parent of $l$
        \State $p.\text{value} \gets \max \{ p.\text{value}, l.\text{value} \}$
        \State $l \gets p$
    \EndWhile
\EndFor
\EndProcedure
\State {\bf end data structure}
\end{algorithmic}
\end{algorithm}

\begin{algorithm}[!ht]\caption{Asynchronous tree data structure, formal version of Algorithm ~\ref{alg:at_query_informal}}\label{alg:at_query}
\begin{algorithmic}[1]
\State {\bf data structure} \textsc{AsynchronousTree}
\State {\bf public:}
\Procedure{Query}{$i \in [n], \tau \in \R_{\geq 0}$} \Comment{\textsc{Query} in  Theorem~\ref{thm:correlation_tree_data_structure}} 
\State \Return \textsc{QuerySub}($\mathrm{root}(T_i),\tau$)
\EndProcedure
\State  
\State {\bf private:}  
\Procedure{QuerySub}{$r\in T, \tau \in \R_{\geq 0}$}
\If{$r$ is leaf}
\If{$r.\text{value}>\tau$}
\State \Return $r$
\EndIf
\Else
\State $r_1\gets$ left child of $r$, $r_2\gets$ right child of $r$
\If{$r_1.\text{value} > \tau$}
    \State $S_1 \gets $\textsc{QuerySub}$(r_1, \tau)$
\EndIf
\If{$r_2.\text{value} > \tau$}
    \State $S_2 \gets $\textsc{QuerySub}$(r_2, \tau)$
\EndIf
\EndIf
\State \Return $S_1 \cup S_2$
\EndProcedure
\State {\bf end data structure}
\end{algorithmic}
\end{algorithm}

\end{document}